\documentclass[sn-mathphys,Numbered]{sn-jnl}

\usepackage{multirow}
\usepackage{booktabs}
\usepackage{amsthm}
\usepackage{mathrsfs}
\usepackage[title]{appendix}
\usepackage{xcolor}
\usepackage{textcomp}
\usepackage{manyfoot}
\usepackage{algorithm}
\usepackage{algorithmicx}
\usepackage{algpseudocode}
\usepackage{listings}
\usepackage{rotating}
\usepackage{array}
\usepackage{graphicx}
\usepackage{hyperref}
\usepackage{url}
\usepackage{amsfonts}
\usepackage{nicefrac}
\usepackage{microtype}
\usepackage{dirtytalk}
\usepackage{float}
\usepackage{caption}
\usepackage{siunitx}
\usepackage{mathtools}
\usepackage{adjustbox}
\usepackage{lineno}
\usepackage{placeins}

\theoremstyle{thmstyleone}%

\theoremstyle{thmstyletwo}%

\theoremstyle{thmstylethree}%
\newcommand{\Frag}[1]{F_{#1}}                 
\newcommand{\FIM}[1]{I_{#1}(\theta)}          

\begin{document}


\title[PICSC]{PIcsC: Partitioning-Induced Covariate Shift Correction }

\author[1]{\fnm{Behraj} \sur{Khan}}\email{behrajkhan@gmail.com}
\author[2]{\fnm{Behroz} \sur{Mirza}}\email{behroz\_mirza@yahoo.com}
\author[3]{\fnm{Syed Ahmad Chan Bukhari} }\email{bukharis@stjohns.edu}
\author*[1]{\fnm{Tahir Qasim} \sur{Syed}} \email{tahirqsyed@gmail.com}

\affil*[1]{\orgdiv{School of Mathematics and Computer Science}, \orgname{Institute of Business Administration Karachi}, \orgaddress{\country{Pakistan}}}

\affil[2]{\orgdiv{Department of Computer Science}, \orgname{Habib University}, \orgaddress{\city{Karachi}, \country{Pakistan}}}

\affil[3]{\orgdiv{Division of Computer Science, Mathematics and Science}, \orgname{St. John's University}, \orgaddress{\country{USA}}}

\abstract{
Covariate shift across training-data partitions biases model selection and parameter estimation in cross-validation, lifelong learning, and federated learning. We propose \textit{Partition-Induced Covariate-shift Correction} (\texttt{PIcsC}), a Fisher information-based regularization framework that mitigates distribution mismatch between data partitions and a reference distribution. \texttt{PIcsC} approximates partition divergence using the Fisher Information Matrix (FIM) and incorporates the resulting statistic as a regularizer during optimization. The same formulation applies to both centrally partitioned datasets (batches or cross-validation folds) and inherently distributed data (federated clients or decentralized nodes), requiring only partition-local gradient statistics rather than raw data. We further introduce a conditional adaptation mechanism that combines FIM shift with KL divergence to detect significant distribution shifts and activates regularization only when necessary. Experiments on more than 40 datasets demonstrate consistent improvements under both natural and synthetic covariate shift. On fragmented batch and fold settings, \texttt{PIcsC} reduces fragmentation-induced performance degradation by more than 20\% and 25\%, respectively. On seven federated learning benchmarks, it consistently outperforms FedAvg, FedProx, and SCAFFOLD by 3 -5 percentage points without requiring client-specific personalization. These results demonstrate that Fisher information provides an effective and unified mechanism for mitigating partition-induced covariate shift across both centralized and distributed learning.
}


\maketitle
\footnote{This work is under consideration in Machine Learning Journal.}
\section{Positioning}
\label{intro}
\noindent
\textbf{Learning systems and feature evolution.}
Big data often demand large learning systems that do not presuppose the presence of the complete dataset in one place at the same time. A contemporary real-world setting therefore undermines the classical machine-learning assumption that data are independently and identically distributed (\textit{iid}), be it among fragments of training data or between training and validation sets. Such problems are characterized as distribution shift \cite{quinonero2008dataset}. The foremost among them is \textit{covariate shift} \cite{cortes2008sample}, variously referred to as \textit{sample selection bias}.

\noindent
\textbf{Dataset fragmentation and covariate shift.}
In covariate shift, the distribution of the covariates (features) changes between training and test time, i.e. $P_{tr}(x) \neq P_{tst}(x)$, while the conditional distribution is unchanged, $P_{tr}(y\mid x) = P_{tst}(y\mid x)$ \cite{sugiyama2007covariate, bickel2007discriminative}. For example, in advertising systems the covariate distribution changes owing to temporal changes in user interest. The risk of covariate shift is especially acute in high-stakes settings such as medical diagnosis \cite{kukar2003transductive}, criminal justice \cite{rudin2018optimized}, and financial analysis \cite{wong2020insights}, and, as a general rule, degrades the performance of any learning system \cite{rezaei2021robust}.

\noindent While the above literature addresses the classical (henceforth \textit{natural}) notion of covariate shift, we are interested in a setting where the \emph{non-colocation} of training data induces covariate shift, because a learner is exposed to several training mini-sets, each with its own empirical distribution, rather than to the pooled dataset at once. Non-colocation arises for two structurally different reasons, both of which we treat under a single formulation:

\begin{enumerate}
    \item[(i)] \textbf{Batch/fold fragmentation.} A single owner deliberately fragments a centrally held dataset into batches \cite{sugiyama2007direct} or folds \cite{moreno2012study} for cross-validation, e.g. in lifelong learning, meta-learning, and online learning \cite{zhang2021adaptive,de2021continual}. Here non-colocation is a design choice: the whole dataset physically exists in one place, but is exposed to the learner piecewise.
    \item[(ii)] \textbf{Client/node fragmentation.} Data is natively partitioned across clients or nodes, as in distributed and federated learning \cite{mcmahan2017communication}, owing to communication cost, ownership, or privacy constraints \cite{gupta2022fl}. Here $X = \bigcup_i \mathcal{D}_i$ is a \emph{virtual} union of client shards $\mathcal{D}_i$ that is never materialized centrally, and statistical heterogeneity across clients (geography, demographics, device, usage pattern) remains a central open problem in federated optimization \cite{solans2024non,ye2023heterogeneous}.
\end{enumerate}

\noindent We index either kind of non-colocated data unit uniformly as a \emph{fragment} $\Frag{i}$, so that in instantiation (i) $\Frag{i} \equiv B_i$ (a batch) or $\Frag{i} \equiv k_i$ (a fold), and in instantiation (ii) $\Frag{i} \equiv \mathcal{D}_i$ (a client/node shard); we use this single symbol $\Frag{i}$ for every statement that holds regardless of instantiation, and switch to the concrete symbol ($B_i$, $k_i$, or $\mathcal{D}_i$) only when a result or table is specific to one setting. Formally, in both (i) and (ii) the learner is exposed to a set of fragments $\{\Frag{1}, \ldots, \Frag{k}\}$, each inducing its own empirical covariate distribution $P_{\Frag{i}}(X)$, such that $P_{\Frag{i}}(X) \neq P_{\Frag{j}}(X)$ for $i \neq j$, even when the fragments are drawn from a single stationary population. We call this \textit{Partition-induced covariate shift (PIcs)}. To the best of our knowledge, covariate shift induced purely by the act of fragmenting  rather than by any change in the underlying population  is uncharted territory in either instantiation, and the two are typically studied by disjoint literatures: cross-validation methodology addresses (i) \cite{moreno2012study}, while federated-optimization methods addressing (ii), such as variance-reduced aggregation \cite{karimireddy2020scaffold} or proximal regularization \cite{li2020federated}, control optimization variance across clients without explicitly modeling or correcting the distributional divergence between them. We argue that (i) and (ii) are two instantiations of the same underlying phenomenon and are amenable to the same remediation.

\noindent
\textbf{Leveraging divergences between distributions.}
Measuring the amount of shift between the distributions of different fragments requires a metric on the space of probability distributions. A natural choice is the relative entropy, or Kullback-Leibler (KL) divergence, used here as a quasi-metric because of its theoretical proximity to the cross-entropy loss of the archetypal classification network. The usual mean-field formulation of the KL divergence uses a Gaussian distribution as the variational prior $q(\theta)$, parametrized by the covariance matrix of the network parameters. This term involves the Hessian of the parameters, whose high dimensionality is a barrier to tractability. Recent inference literature \cite{pascanu2013revisiting, bukaew2021one} approximates the second-order derivative of the KL divergence by the Fisher Information Matrix (FIM), which can itself be estimated from the variance and expectation of the network's parameter gradients \cite{nishiyama2019new}. Penalizing a function of the FIM between a fragment and a reference distribution therefore serves as an antidote to partition-induced covariate shift whether between a batch/fold and the validation set, or between one client's local update and the accumulated prior of previously seen clients. Because this estimator requires only fragment-local gradient statistics (Section~\ref{sec:two-instantiations}) rather than raw data, it is directly compatible with the non-colocation constraint that motivates instantiation (ii). In the paper, we play on the letter similarity between the problem (\textit{FI}cs) and the solution (\textit{FI}M) for nomenclature, and additionally read PIcsC as \say{Pisces} \footnote{The acronym PICSC is deliberately close in sound to \emph{Pisces}, Latin for \say{fish}. The pun is not merely decorative: a fish maintains its trajectory only by continuously sensing and correcting for the shifting currents around it, rather than assuming the water is still. Covariate distributions across data partitions behave analogously  rarely stationary, and prone to shift whether across batches, folds, or federated clients (Section~\ref{intro})  so that a learner, like the fish, must continually re-estimate and correct its course rather than assume the distribution it was first trained on still holds.}.

\noindent
\textbf{Contributions.}
The contributions of the paper are:
\begin{enumerate}
  \item We identify and remediate covariate distribution shift when fragments of data  batches, folds, or client/node shards  are separately cross-validated or separately trained on.
  \item We give a computationally tractable estimator of the divergence between a fragment's parameter posterior and a reference, using only that fragment; memory cost is linear in the size of one fragment rather than the complete dataset, and the estimator requires only fragment-local gradient summaries, not raw data.
  \item We introduce an information-theoretic penalty that accumulates this information across previously seen fragments and regularizes the parameters learned from the current fragment.
  \item We show this Fisher-KL estimator is agnostic to the physical or organizational source of fragmentation: the identical estimator applies whether fragments arise from a chosen batching/fold schedule over centrally held data, or from data natively non-colocated across federated clients/nodes.
  \item We extend $PIcsC$ from an unconditional, all-pairs penalty to a \emph{conditional} detection-and-adaptation mechanism  a composite Fisher-KL trigger $\tau_t$ coupled with a temporally smoothed, Cram\'er-Rao-anchored regularizer  enabling constant-memory, boundary-free operation over long batch or client sequences, and we validate it on seven federated benchmarks against FedAvg, FedProx, and SCAFFOLD.
\end{enumerate}

\noindent
\textbf{Paper organization.}
The remainder of the paper is organized as follows. Section~\ref{sec:method} develops the PIcsC estimator: it introduces the unified fragment notation used throughout (Section~\ref{sec:two-instantiations}), derives the Fisher-based approximation to the parameter divergence and the resulting penalized loss (Section~\ref{sec:fim-approx}), establishes the convexity of the FIM penalty (Section~\ref{sec:convexity}), and extends the unconditional penalty to a conditional detection-and-adaptation mechanism suited to long or federated fragment sequences (Section~\ref{sec:conditional}). Section~\ref{bnchmrks} situates PIcsC relative to existing covariate-shift correction methods and federated-optimization baselines. Section~\ref{3} describes the experimental setup, including datasets, model architectures, baselines, and the design of experiments E1-E6, spanning both the batch/fold and client/node instantiations. Section~\ref{4} reports and discusses results for E1-E6, an ablation study, and a synthesis across both instantiations. Section~\ref{5} concludes.

\section{Method}
\label{sec:method}
We require a quantification of the amount of distribution shift between a pair of fragments most commonly a batch/fold and the validation set, but possibly more generally fragment against fragment. This quantity is added as a corrective term to the loss computed on the following fragment(s), so that the loss remains close to the conventional cross-validation baseline. The penalty accrues over processed fragments and gives a running estimate of PIcs, which becomes the \textit{parameter prior} used to compute the PIcs of the next fragment.

\subsection{Notation and the fragment sequence}
\label{sec:two-instantiations}
Let $x \sim X$ be a training example drawn from a complete training dataset $X$ (materialized centrally in the batch/fold instantiation, or virtual and never pooled in the client instantiation). $X$ is fragmented into a set of $k$ fragments, denoted $\{\Frag{i}\}_{i=1}^{k} = \{\Frag{1}, \Frag{2}, \ldots, \Frag{k}\}$, indexed by $i \in \mathcal{I} = \{1, 2, \ldots, k\}$. The index set $\mathcal{I}$ carries different meaning under the two instantiations of Section~\ref{sec:two-instantiations}: in the batch/fold instantiation, $i$ is a strictly increasing temporal index, so that $\Frag{i}$ precedes $\Frag{i+1}$ in the order in which fragments are presented to the learner; in the client/node instantiation, $i$ is an arbitrary but fixed enumeration of clients/nodes, with no implied temporal order between $\Frag{i}$ and $\Frag{j}$ for $i \neq j$. We use $\{\Frag{i}\}_{i=1}^k$ (rather than dot notation) whenever the distinction between an ordered sequence and an unordered index set matters, and revert to the shorthand $\{\Frag{1},\ldots,\Frag{k}\}$ elsewhere for readability. $X$ follows a training data distribution $P(X)$, and, given a parametrized learning algorithm, there is an induced parameter distribution: $P(\theta_{\Frag{i}})$ denotes the parameters obtained after training on fragment $\Frag{i}$. We write $P(\theta)$ for a generic parameter distribution when the fragment index is not material. $f_\theta(x)$ denotes the model's fit to the training data, regardless of fragmentation. Two concrete instantiations of $\Frag{i}$ recur throughout the paper:

\begin{itemize}
    \item \textbf{Batch/fold instantiation.} $\Frag{i} = B_i$ (a batch) or $\Frag{i} = k_i$ (a fold), obtained by partitioning a centrally available dataset $X$ under a chosen splitting ratio (Section~\ref{3}). The index $i$ reflects the order in which fragments are processed for cross-validation and is a modelling choice.
    \item \textbf{Client/node instantiation.} $\Frag{i} = \mathcal{D}_i$, the local dataset held by client or node $i$ in a distributed or federated system, with $X = \bigcup_i \mathcal{D}_i$ a virtual union that is never centrally materialized. Consistent with the definition of $\mathcal{I}$ above, the index $i$ here is a fixed enumeration (not a temporal order); where communication rounds are relevant (Section~\ref{sec:conditional}), we index rounds separately by $t$.
\end{itemize}

\noindent Given a target distribution $Q(y \mid x)$ for input $x$, the KL divergence from source $P$ to target $Q$ is
\[
D_{KL}\big(P(y\mid x) \,\big\|\, Q(y \mid x)\big) = \sum_y P(y \mid x) \log\!\left(\frac{P(y\mid x)}{Q(y\mid x)}\right)
\]
treating $y$ as a continuous target variable and $P(y\mid x)$, $Q(y\mid x)$ as an arbitrary and a Gaussian distribution respectively, with means $\mu_P, \mu_Q$ and variances $\sigma_P^2, \sigma_Q^2$. Everything that follows is stated for the abstract fragment sequence $\{\Frag{1},\ldots,\Frag{k}\}$; results and algorithms are agnostic to which of the two instantiations produced it, and we instantiate concretely (as $B_i$/$k_i$ or $\mathcal{D}_i$) only in the experimental sections.

\subsection{Approximating the parameter divergence with the Fisher Information Matrix}
\label{sec:fim-approx}
Consider a model with parameters $\theta$ and likelihood $p(X \mid \theta)$ for observed data $X$, and let $\hat\theta$ be an estimator of the true parameter $\theta$. The Fisher information matrix is the expected negative Hessian of the log-likelihood,
\begin{equation}
I(\theta) = \mathbb{E}\left[-\frac{\partial^2 \log p(X \mid \theta)}{\partial \theta\, \partial \theta^\top}\right].
\label{eq:fim-def}
\end{equation}
The Cram\'er-Rao lower bound (CRLB) states that, for any unbiased estimator $\hat\theta$, the covariance $V(\hat\theta)$ satisfies
\begin{equation}
V(\hat\theta) \succeq I^{-1}(\theta),
\label{eq:crlb}
\end{equation}
where $\succeq$ denotes that $V(\hat\theta) - I^{-1}(\theta)$ is positive semi-definite. Approximating the variational posterior $q(\hat\theta)$ around $\theta$ by a Gaussian with mean $\theta$ and covariance $V(\hat\theta)$,
\begin{equation}
q(\hat\theta) \approx \mathcal{N}\big(\theta, V(\hat\theta)\big),
\label{eq:gaussian-var}
\end{equation}
the KL divergence between a fragment's parameter posterior $p(\theta)$ and this Gaussian approximation can be written
\begin{equation}
D_{KL}\big(p(\theta) \,\|\, q(\hat\theta)\big) \approx \int p(\theta) \log\!\left(\frac{p(\theta)}{\mathcal{N}(\theta, V(\hat\theta))}\right) d\theta.
\label{eq:dkl-param}
\end{equation}
Substituting the CRLB \eqref{eq:crlb}, $V(\hat\theta) \succeq I^{-1}(\theta)$, into \eqref{eq:dkl-param} gives the working approximation used throughout the paper,
\begin{equation}
D_{KL}\big(p(\theta) \,\|\, q(\hat\theta)\big) \approx \int p(\theta) \log\!\left(\frac{p(\theta)}{\mathcal{N}(\theta, I^{-1}(\theta))}\right) d\theta,
\label{eq:dkl-fisher}
\end{equation}
i.e. the relative entropy between a fragment's parameter distribution and a reference is estimated from the variance-covariance structure of the estimated parameters, via the FIM. An unbiased estimator $\hat\theta(X_1,\ldots,X_n)$ of $\theta$ over $n$ samples further satisfies the sample form of the CRLB,
\begin{equation}
\sigma^2(\hat\theta) \geq \frac{1}{n\, I(\theta)}, \qquad I(\theta) = -\mathbb{E}\left[\frac{\partial^2 \log P(X;\theta)}{\partial \theta^2}\right],
\label{eq:crlb-sample}
\end{equation}
which we use in Section~\ref{sec:two-instantiations} and Section~\ref{sec:conditional} to justify estimating $I(\theta)$ locally, from the gradients available on a single fragment, without access to any other fragment or to raw examples from it. In the batch/fold instantiation, $I(\theta)$ is estimated on the current batch/fold against the validation set; in the client/node instantiation, $I(\theta)$ is estimated locally at each client (Section~\ref{sec:two-instantiations}), so that a client need only transmit its (diagonal) Fisher summary and current parameters to the aggregator, never its raw data  preserving the privacy property that motivates non-colocation in the first place, and bounding per-client communication and memory to $\mathcal{O}(d)$ in the parameter count $d$.\\

\noindent An important result \cite{courtade2016monotonicity} shows the monotonicity of entropy and Fisher information, i.e. the latter remains strictly upper-bounded by the former; we use this to justify replacing the full divergence term of \eqref{eq:dkl-fisher} by the FIM alone at a significant reduction in computation. The resulting penalized loss for a single fragment is
\begin{equation}
\mathcal{L}(x,y;\theta) = -\int P(y \mid x) \log\big(P(y\mid x;\theta)\big)\, d\theta \;-\; \lambda \int \frac{\partial^2 \log p(X \mid \theta)}{\partial \theta\, \partial\theta^\top}\, d\theta,
\label{eq:penalized-loss}
\end{equation}
where $\lambda$ is the penalty strength calibrated in Section~\ref{cp}. Algorithm~\ref{algo1} gives the resulting fragmented cross-validation procedure, applying the penalty of \eqref{eq:penalized-loss} unconditionally between every pair of fragments.

\subsection{The FIM penalty is convex}
\label{sec:convexity}
It is important to relate the FIM to $D_{KL}$. For source and target distributions $P(x)$, $Q(x)$ with common support,
\begin{equation}
D_{KL}(P\|Q) = \int_X P(x) \log\!\left(\frac{Q(x)}{P(x)}\right) dx.
\label{eq:kl-integral}
\end{equation}
Let $P(x;\theta)$ be the true distribution for input $X$ with parameter $\theta$, and $Q(x;\hat\theta)$ an arbitrary target distribution with parameter $\hat\theta$; then
\begin{equation}
D_{KL}\big(P(\cdot;\theta)\,\|\,Q(\cdot;\hat\theta)\big) = \mathbb{E}_{X\sim P(\cdot;\theta)}\left[\log\!\left(\frac{Q(X;\hat\theta)}{P(X;\theta)}\right)\right].
\label{eq:dkl-parameterized}
\end{equation}
For the special case $Q(x;\hat\theta) = P(x;\hat\theta)$, minimizing $D_{KL}$ with respect to $\hat\theta$ attains its minimum at $\hat\theta = \theta$, so
\begin{equation}
D\big(P(\cdot;\theta)\,\|\,P(\cdot;\hat\theta)\big) \geq 0.
\label{eq:dkl-nonneg-def}
\end{equation}
A second-order Taylor expansion of $\log P(x;\hat\theta)$ around $\theta$ gives
\begin{align}
\log P(X;\hat\theta) &= \log P(X;\theta) + (\hat\theta-\theta)\frac{\partial \log P(X;\theta)}{\partial \theta} \nonumber\\
&\quad - \tfrac{1}{2}(\hat\theta-\theta)^2 \frac{\partial^2 \log P(X;\theta)}{\partial \theta^2} + O\big((\hat\theta-\theta)^3\big).
\label{eq:taylor}
\end{align}
Taking the expectation with respect to $X \sim P(\cdot;\theta)$,
\begin{align}
\mathbb{E}_{X\sim P(\cdot;\theta)}\big[\log P(X;\hat\theta) - \log P(X;\theta)\big] &= (\hat\theta-\theta)\,\mathbb{E}_{X\sim P(\cdot;\theta)}\left[\frac{\partial \log P(X;\theta)}{\partial \theta}\right] \nonumber \\
&\quad - \tfrac{1}{2}(\hat\theta-\theta)^2\, I(\theta) + O\big((\hat\theta-\theta)^3\big).
\label{eq:taylor-expectation}
\end{align}
The left-hand side of \eqref{eq:taylor-expectation} is $-D_{KL}\big(P(\cdot;\theta)\,\|\,P(\cdot;\hat\theta)\big)$, which by \eqref{eq:dkl-nonneg-def} is non-positive; hence the right-hand side is non-positive as well, which, since it holds for arbitrary $\hat\theta$, requires
\begin{equation}
I(\theta) \geq 0.
\label{eq:fim-nonneg}
\end{equation}
The FIM is therefore convex (positive semi-definite), and its addition to the network loss of \eqref{eq:penalized-loss} does not alter the loss's monotonicity properties.

\begin{algorithm}[b]
  \caption{Fragmentation (batchwise, foldwise, or client/node-wise) and partition-induced covariate shift mitigation}
  \label{algo1}
  \begin{algorithmic}[1]
    \State \textbf{Require: } model \( f_\theta \) parameterized by $\theta$;
    \Statex training data \(\mathcal{D}_{tr}\) (centrally held, or the virtual union $\bigcup_i \mathcal{D}_i$ of non-colocated client/node shards);
    \Statex validation data \(\mathcal{D}_{v}\) (or an accumulated fragment prior, in the client/node instantiation);
    \Statex \hspace*{2.2em} number of fragments \(K\);
    \Statex \hspace*{2.2em} number of epochs \(T\);
    \Statex  $\mathcal{D}_i$ denotes fragment $\Frag{i}$'s local data (a batch, fold, or client/node shard, per Section~\ref{sec:two-instantiations}), for $i = 1, \ldots, K$
   \Procedure{ShiftRemediation}{$\mathcal{D}_{tr}$, $\mathcal{D}_{v}$}
      \State \textbf{obtain} the $K$ fragments $\{\Frag{1},\ldots,\Frag{K}\}$  by splitting \(\mathcal{D}_{tr}\) into batches/folds, or by enumerating the $K$ client/node shards
      \State \textbf{initialize} \( f_\theta \)
      \For{$\text{epoch} \gets 1$ \textbf{to} $T$}
        \State $\mathcal{L}_{\text{total}} \leftarrow 0$
        \For{$i \gets 1$ \textbf{to} $K$}
          \For{$j \gets i+1$ \textbf{to} $K$}
            \State $d_{ij} \leftarrow D_{KL}\big(P(\mathcal{D}_i) \,\|\, P(\mathcal{D}_j)\big)$ \Comment{FIM-based estimate on fragments $\mathcal{D}_i, \mathcal{D}_j$, Eqs.~\eqref{eq:fim-def}-\eqref{eq:crlb-sample}}
            \State $\mathcal{L}_{\text{total}} \leftarrow \mathcal{L}_{\text{total}} + \mathcal{L}(x,y;\theta) + \lambda\, d_{ij}$ \Comment{Eq.~\eqref{eq:penalized-loss}: task loss plus $\lambda$-weighted penalty $d_{ij}$}
          \EndFor
        \EndFor
        \State update \( f_\theta \) using \(\mathcal{L}_{\text{total}}\) \Comment{single gradient step on the accumulated, penalized loss}
      \EndFor
      \State \textbf{return} \( f_\theta \)
    \EndProcedure
  \end{algorithmic}
\end{algorithm}

\begin{algorithm}[b]
\caption{Conditional $PIcsC$: detection and adaptation over a fragment sequence (batches, folds, or clients)}
\label{algo2}
\begin{algorithmic}[1]
\Require Fragment sequence $\{\Frag{1}, \ldots, \Frag{k}\}$, initial parameters $\theta_0$, smoothing rate $\alpha$, regularization strength $\lambda$, detection threshold $\gamma$
\State Initialize $I_{\mathrm{global}} \leftarrow \epsilon I_d$
\For{$t = 1$ \textbf{to} $k$}
    \State Compute diagonal $\FIM{t-1}$ on $\Frag{t}$ (parameters $\theta_{t-1}$)
    \State $\tau_t \leftarrow \big\|I_t - I_{t-1}\big\|_F \cdot D_{KL}\big(P_{\Frag{t}}\,\|\,P_{\Frag{t-1}}\big)$ \Comment{Eq.~\eqref{eq:tau}}
    \If{$\tau_t > \gamma$}
        \State $I_{\mathrm{global}} \leftarrow \alpha\, I_{\mathrm{global}} + (1-\alpha)\, I_t$ \Comment{Eq.~\eqref{eq:fim-smooth}}
        \State $\theta_t \leftarrow \arg\min_\theta \mathcal{L}_t(\theta)$ \Comment{Eq.~\eqref{eq:loss-conditional}}
    \Else
        \State $\theta_t \leftarrow \theta_{t-1}$
    \EndIf
\EndFor
\end{algorithmic}
\end{algorithm}
\subsection{From an unconditional penalty to conditional detection-and-adaptation}
\label{sec:conditional}
Algorithm~\ref{algo1} applies the FIM-based penalty at every pair of fragments unconditionally, which is tractable when $k$ is small and bounded, as in the batch/fold instantiation (Section~\ref{3}). For long or unbounded fragment sequences  in particular streaming and federated deployments, where new clients or rounds arrive indefinitely. We propose a \emph{conditional} variant that detects when remediation is warranted before applying it, giving constant per-fragment memory and avoiding unnecessary regularization when consecutive fragments are not meaningfully shifted.

\noindent We define a composite shift signal between consecutive fragments $\Frag{t-1}$ and $\Frag{t}$, coupling geometric and distributional change:
\begin{equation}
\tau_t = \big\|\FIM{t} - \FIM{t-1}\big\|_F \cdot D_{KL}\big(P_{\Frag{t}} \,\|\, P_{\Frag{t-1}}\big),
\label{eq:tau}
\end{equation}
where $\FIM{t}$ is the FIM estimated on fragment $\Frag{t}$ (Eq.~\eqref{eq:fim-def}) and $\|\cdot\|_F$ is the Frobenius norm. $D_{KL}$ alone is a purely distributional signal that can be uninformative in high dimensions, where many density changes leave a downstream model's loss landscape largely unaffected; the Frobenius shift of the FIM alone lacks a probabilistic interpretation, and can be misled by curvature changes that do not reflect an actual distributional shift (e.g. changes induced solely by optimization dynamics within a stationary fragment). The product form of \eqref{eq:tau} suppresses false positives that either term would raise in isolation. A shift is flagged when $\tau_t > \gamma$, for a calibrated threshold $\gamma$.

When flagged, we update a temporally smoothed global FIM,
\begin{equation}
I_{\mathrm{global}}^{(t)} = \alpha\, I_{\mathrm{global}}^{(t-1)} + (1-\alpha)\, \FIM{t},
\label{eq:fim-smooth}
\end{equation}
with $\alpha \in [0,1]$ controlling temporal sensitivity, and regularize the objective for fragment $\Frag{t}$ as
\begin{equation}
\mathcal{L}_t(\theta) = \underbrace{\mathbb{E}_{(x,y)\sim P_{\Frag{t}}}\big[\ell(f_\theta(x),y)\big]}_{\text{current-fragment loss}} \;+\; \lambda \underbrace{(\theta-\mu_{t-1})^\top I_{\mathrm{global}}^{(t-1)} (\theta-\mu_{t-1})}_{\text{Cram\'er-Rao regularizer, Eq.~\eqref{eq:crlb}}},
\label{eq:loss-conditional}
\end{equation}
where $\mu_{t-1}$ is the previous parameter estimate. Equation~\eqref{eq:loss-conditional} penalizes movement along directions in which the FIM indicates the model was previously highly sensitive  and hence, by the CRLB \eqref{eq:crlb}, previously estimated with relatively low variance  while leaving low-sensitivity directions free to adapt to the incoming fragment. When $\tau_t \leq \gamma$, no update to $I_{\mathrm{global}}$ or extra regularization is applied, i.e. $\theta_t \leftarrow \theta_{t-1}$.

\noindent In the client/node instantiation, $\Frag{t}$ indexes clients rather than time steps, and Eqs.~\eqref{eq:tau}-\eqref{eq:loss-conditional} are computed identically, with each client transmitting only its local diagonal FIM and parameter update to the aggregator never raw data  preserving the non-colocation constraint. Algorithm~\ref{algo2} gives the conditional procedure; it subsumes Algorithm~\ref{algo1} as the special case $\gamma = 0$ (always trigger).

\section{Benchmarks and state-of-the-art}
\label{bnchmrks}

Few published works address covariate-shift mitigation directly under fragmentation. Most existing work instead relies on weight estimation via statistical optimization \cite{sugiyama2012density,huang2006correcting,cortes2008sample}. Shimodaira~\cite{shimodaira2000improving} assigns an importance weight (IW) to each training example to reduce variance and balance the bias-variance trade-off under covariate shift. Sugiyama et al.~\cite{sugiyama2007covariate} address model selection under natural covariate shift using importance-weighted cross-validation (IW-CV). Importance-based approaches perform well in classical (non-deep) settings but degrade in high dimensions. Fang et al.~\cite{fang2020rethinking} propose dynamic importance weighting (DIW) for deep models under dataset shift, iterating weighted estimation and weighted classification to update the feature extractor and classifier jointly; DIW requires a validation set (from train or test) to pretrain the classifier, which constrains its applicability. Other work minimizes various divergences between source and target distributions \cite{sugiyama2012density,tsuboi2009direct,stojanov2019low,zhang2023adapting,sugiyama2012machine}. Gretton et al.~\cite{gretton2009covariate} use kernel mean matching (KMM), matching the mean embeddings of source and target in a reproducing kernel Hilbert space; KMM assumes an unbiased density ratio between train and test, an assumption that fails under genuine covariate shift. Moreno-Torres et al.~\cite{moreno2012study} study the shift induced by $k$-fold cross-validation and propose cross-validation variants to adapt to it. All of these approaches assume the dataset resides in memory in its entirety; to our knowledge, none addresses the setting where fragments of data are shown to a network sequentially or across nodes and must be reconciled without ever being pooled.\\

\noindent A further body of related work studies distributional shift in task-agnostic continual and lifelong learning, where, as in our setting, no explicit task boundaries are available to trigger adaptation. \cite{faber2023vlad} study covariate shift detection in continual learning without task boundaries; \cite{basterrech2022tracking} address distribution shift in streaming/lifelong settings using online statistical monitoring; and \cite{cano2022rose} examine related aspects of concept and covariate shift under task-agnostic continual learning. These works share our premise that distributional change can arise purely from how data is sequentially exposed to a learner, independent of any change in the underlying task; they differ from $PIcsC$ in that they target within-stream shift \emph{detection} for a single continual learner, whereas $PIcsC$ additionally derives a Fisher-Cram\'er-Rao regularizer that corrects for the shift once detected, and treats batch/fold and federated-client fragmentation as instances of the same underlying phenomenon (Section~\ref{intro}).\\

\noindent A parallel literature addresses client heterogeneity in federated learning through optimization-side fixes rather than explicit distributional correction: FedAvg \cite{mcmahan2017communication} aggregates client updates under an implicit assumption of comparably distributed clients; FedProx \cite{li2020federated} adds a proximal term for local-update stability; SCAFFOLD \cite{karimireddy2020scaffold} uses control variates to reduce client-shift variance. None of these explicitly quantifies or corrects the covariate shift between clients in the sense of Section~\ref{intro}. Because our Fisher-KL estimator (Eq.~\eqref{eq:fim-def}) is agnostic to whether fragments are batches/folds or clients (Section~\ref{sec:two-instantiations}), it is complementary to, rather than competing with, this line of work; we validate this directly, using the conditional mechanism of Section~\ref{sec:conditional}, against these three federated baselines in Section~\ref{sec:fl-results}.

\section{Experimental Setup}
\label{3}

We demonstrate the efficacy of the proposed method against multiple baselines for partition-induced covariate shift, and on benchmarks for natural covariate shift used as a surrogate. This section describes, in turn, the datasets (Section~\ref{datasets}), model architectures (Section~\ref{sec:models}), baselines (Section~\ref{sec:baselines}), experimental design (Section~\ref{sec:expdesign}), and implementation details (Section~\ref{sec:impl}), covering both the batch/fold (temporal) and client/node (federated, spatial) instantiations of Section~\ref{sec:two-instantiations}.

\subsection{Datasets}
\label{datasets}
The proposed method evaluated on 49 datasets including 27 binary, 13 image-based and 9 federated benchmarking datasets. The datasets are given below:

\begin{itemize}
    \item \textbf{Standard vision benchmarks (no pre-existing covariate shift):} MNIST \cite{lecun1998mnist}, Fashion-MNIST \cite{xiao2017fashion}, and CIFAR-10/CIFAR-100 \cite{krizhevsky2009learning}.

    \item \textbf{Vision benchmarks with natural or induced covariate shift:} KMNIST \cite{clanuwat2018deep}, SVHN \cite{Netzer2011}, Caltech101 \cite{FeiFei2004LearningGV}, Tiny-ImageNet, STL-10 \cite{coates2011analysis}, Permuted MNIST (P-MNIST) \cite{goodfellow2013empirical}, MNIST-C \cite{mu2019mnist}, and CIFAR10-C/CIFAR100-C \cite{hendrycks2019robustness}.

    \item \textbf{Tabular benchmarks:} 27 binary-classification datasets from the KEEL repository \cite{alcala2011keel}, accessed through OpenML.

    \item \textbf{Federated learning benchmarks:} FEMNIST \cite{caldas2018leaf}, CIFAR-10, CIFAR-100 \cite{krizhevsky2009learning}, SVHN \cite{Netzer2011}, EMNIST Digits, Amazon Reviews \cite{hou2024bridging}, and Shakespeare \cite{mcmahan2017communication}, using their standard client-based partitions or established non-IID client splits.
\end{itemize}

\noindent Accuracy is used as the evaluation metric throughout, reported per-fragment and averaged across fragments for the batch/fold instantiation (Sections~\ref{4}), and as mean $\pm$ standard deviation across clients for the federated instantiation (Section~\ref{sec:fl-results}).


\subsection{Model architectures}
\label{sec:models}
For all image-based benchmarks we use a five-layer convolutional neural network (CNN) with softmax cross-entropy loss, comprising two convolutional layers with pooling followed by three fully connected layers; this architecture is held fixed across all image datasets. For tabular data we use a multi-layer perceptron with one hidden layer of four units, ReLU activations, and the Adam optimizer; this architecture is held fixed across all 27 KEEL datasets. For the federated benchmarks (Section~\ref{sec:fed-setup}), each client uses the architecture appropriate to its data modality. The CNN described above for the federated image benchmarks, and a sequence/text model appropriate to each corpus for Amazon Reviews and Shakespeare, consistent with standard practice for these benchmarks. The FIM used in Algorithms~\ref{algo1}-\ref{algo2} is approximated by its diagonal, estimated via the empirical-Fisher estimator on the fragment's local data (Section~\ref{sec:fed-setup}).

\subsection{Baselines}
\label{sec:baselines}

We compare $PIcsC$ against the following baselines.

\begin{itemize}
    \item \textbf{Cross-validation baselines:} standard cross-validation on the unfragmented training set, batch-wise fragmented training set, and $k$-fold fragmented training set.

    \item \textbf{Covariate-shift correction methods:} importance weighting (IW) \cite{shimodaira2000improving}, importance-weighted cross-validation (IWCV) \cite{sugiyama2007covariate}, kernel mean matching (KMM) \cite{gretton2009covariate}, and dynamic importance weighting (DIW) \cite{fang2020rethinking}.

    \item \textbf{Federated learning baselines:} FedAvg \cite{mcmahan2017communication}, FedProx \cite{li2020federated}, and SCAFFOLD \cite{karimireddy2020scaffold}, each evaluated under the same client sequence and using the hyperparameter settings recommended in the original publications.
\end{itemize}

\subsection{Experimental design}
\label{sec:expdesign}
We design six experiments spanning both fragmentation instantiations. Throughout, training data is seen in two guises: \textbf{integral} (the unfragmented training set) and \textbf{fragmented} (split for cross-validation, in either the batch/fold or client/node instantiation). Each fragmented training set uses one of five splitting ratios (5\%, 10\%, 20\%, 25\%, 50\%), yielding (20, 10, 5, 4, 2) batches respectively; individual batches are denoted $B_1, B_2, B_3, \ldots, B_{n/2}, \ldots, B_{n-1}, B_n$ for $n$ the number of batches in a given setting.

\begin{itemize}
    \item \textbf{E1:  Evidencing covariate shift induced by batching.} Does fragmenting a dataset into batches induce PIcs? We run standard cross-validation (st-CV) on the integral training set (\textbf{BL1}) and on the fragmented training set (\textbf{BL2}), holding the validation set fixed, on both no-covariate-shift and natural-covariate-shift datasets (Section~\ref{datasets}). Results are reported in Table~\ref{tab: bst-CV-full} (Section~\ref{sec:e1e2-results}).
    \item \textbf{E2:  Mitigating PIcs with $PIcsC$.} Does $PIcsC$ correct the shift observed in E1? We run $PIcsC$ (Algorithm~\ref{algo1}) on the integral and fragmented training sets, on both no-covariate-shift and natural-covariate-shift datasets, and compare against \textbf{BL2}. We further compare the mean accuracy across batches, $\mu_1$ (st-CV, Table~\ref{tab: bst-CV-full}) against $\mu_2$ ($PIcsC$, Table~\ref{tab: ccca}).
    \item \textbf{E3: Evidencing the aggravation of PIcs under $k$-fold CV.} Does $k$-fold cross-validation aggravate PIcs relative to batching? We fragment each dataset into $k \in \{2,5,10\}$ folds, run st-CV foldwise (\textbf{BL3}), and compare against \textbf{BL1}. Results are reported in Tables~\ref{tab: cccakeel} (tabular) and \ref{tab:st-CVfold} (image).
    \item \textbf{E4: Correcting PIcs in the fold setting.} Does $PIcsC$ correct the shift observed in E3? We run $PIcsC$ foldwise for $k \in \{2,5,10\}$ and compare against \textbf{BL3}. \cite{moreno2012study} hypothesize that covariate shift is induced by the $k$-fold procedure itself; unlike the batchwise setting, the foldwise setting has no held-out reference (validation) set against which to measure shift. Results are reported in Tables~\ref{tab: cccakeel}, \ref{tab: c3keel} (tabular) and \ref{tab:st-CVfold}, \ref{tab: c3fold} (image).
    \item \textbf{E5: Positioning $PIcsC$ against the state of the art.} We run $PIcsC$ on the integral training set for all no-covariate-shift and natural-covariate-shift datasets and compare against the benchmarks of Section~\ref{bnchmrks} and \textbf{BL1}. Results are reported in Table~\ref{tab: meanacc}.
    \item \textbf{E6: Client/node fragmentation (federated instantiation).} Does client/node fragmentation induce PIcs, and does the conditional mechanism of Section~\ref{sec:conditional} correct it, analogously to E1/E2 for batches? Each client trains locally, transmits its diagonal FIM and parameter update (never raw data) to the aggregator, and the server applies Algorithm~\ref{algo2} whenever $\tau_t > \gamma$. We compare against FedAvg, FedProx, and SCAFFOLD (Section~\ref{sec:baselines}) on the seven federated benchmarks of Section~\ref{datasets}. Results are reported in Table~\ref{tab: fedresults}.
\end{itemize}

\noindent An ablation study (Section~\ref{sec:ablation}) additionally examines robustness to varying amounts of injected covariate shift, and to the number of fragments, on CIFAR-10 and CIFAR-100.

\subsection{Implementation details}
\label{sec:impl}
\textbf{Hyperparameters.} We set the penalty strength $\lambda \in \{0.01, 0.04, 0.07, 0.1\}$ across preliminary experiments; $PIcsC$ performs best at $\lambda = 0.1$ (Section~\ref{cp}), and we report all main results at this value. For the conditional mechanism (Section~\ref{sec:conditional}), the FIM smoothing rate $\alpha$ and detection threshold $\gamma$ are additionally selected on a held-out validation stream per federated dataset.\\

\noindent The calibration in Figure~\ref{fig:lambda} is conducted across the fold-wise KEEL (tabular, MLP) settings; we adopt the resulting $\lambda=0.1$ uniformly for the CNN-based image experiments as well, rather than recalibrating $\lambda$ per architecture, and this is a simplification rather than a claim that $\lambda=0.1$ is architecture- or loss-invariant in general. Because the penalty of Eq.~\eqref{eq:penalized-loss} scales with the FIM, which is itself sensitive to a model's parameter count, loss curvature, and output dimensionality (e.g. binary versus multi-class cross-entropy), we would expect the optimal $\lambda$ to shift somewhat between the MLP and CNN architectures used here, and more so for architectures not evaluated in this paper (e.g. transformers). We did not observe evidence of this in practice (the same $\lambda$ performed well across both settings we tested), but we have not conducted a dedicated sensitivity sweep of $\lambda$ per architecture/loss combination, and we flag a systematic study of this sensitivity, together with a possible adaptive or per-layer $\lambda$ schedule, as a direction for future work (Section~\ref{sec:discussion}).\\

\noindent\textbf{Software and hardware.} All baselines are implemented in TensorFlow 2.11\footnote{\url{www.tensorflow.org}}; code is available at an anonymized repository.\footnote{\url{https://github.com/behraj/PIcsC}} We reproduce baseline results as published where available. All experiments are run on an RTX 3090 Ti GPU with 24\,GB of GPU memory and 128\,GB of system memory. Average runtime is 30 minutes per experiment on the integral training set and 210 minutes on the fragmented training set; runtime varies with dataset size and the number of batch/fold splits.\\

\noindent\textbf{FIM estimator and cost.} Throughout, $I(\theta)$ (Eq.~\eqref{eq:fim-def}) is approximated by the \emph{diagonal empirical Fisher}: the squared per-parameter gradient of the log-likelihood, averaged over a fragment's local examples, rather than the exact Hessian, the full (non-diagonal) Fisher, or an inverse Gram/NTK-style matrix. This is a standard approximation in Fisher-regularized continual learning (e.g. EWC-style methods) and avoids ever forming or inverting a $d\times d$ matrix, where $d$ is the parameter count. Concretely, the diagonal empirical Fisher requires one additional backward pass per fragment beyond the standard forward/backward pass already needed for the task loss (squaring and averaging per-parameter gradients already computed for $\mathcal{L}(x,y;\theta)$), so its marginal FLOP cost is within a small constant factor of one extra gradient computation, and its memory cost is $\mathcal{O}(d)$  the size of a single parameter vector  rather than $\mathcal{O}(d^2)$ for a full or Gram-matrix FIM. This is what makes the method practical for the CNN and MLP architectures used throughout (Section~\ref{sec:models}) on standard GPU hardware (Section~\ref{sec:impl}): the added cost is comparable to computing one extra gradient per fragment, not to a second-order (Hessian) optimization step.

\section{Results and Discussion}
\label{4}
This section reports and discusses results for experiments E1-E6 (Section~\ref{sec:expdesign}), followed by an ablation study and a synthesis of the findings across both fragmentation instantiations.

\subsection{E1/E2: Covariate shift induced by batching, and its mitigation}
\label{sec:e1e2-results}

\textbf{No-covariate-shift datasets.} Table~\ref{tab: bst-CV-full} shows that fragmenting data decreases both average and per-batch accuracy relative to the st-CV baseline (BL1), consistent with covariate shift induced by the fragmentation operation itself: average accuracy falls by more than $36\%$ across the $20$-, $10$-, and $2$-batch fragmentations.

\noindent \textbf{Natural-covariate-shift datasets.} The same pattern holds, but with a markedly steeper decline: Table~\ref{tab: bst-CV-full} shows average accuracy falling by more than $60\%$ across the $20$-, $10$-, and $2$-batch fragmentations relative to BL1. This sharper drop is consistent with a compounding effect, whereby partition-induced shift is superimposed on the natural shift already present in these datasets.

\noindent \textbf{Correlation with fragmentation frequency.} Figure~\ref{fig:delta} and Table~\ref{tab: bst-CV-full} show a positive relationship between the decrease in average accuracy and the number of batches: on the no-covariate-shift datasets, the drop moderates from $52.1\%$ over $20$ batches to $43.7\%$ over $10$ batches and $36.3\%$ over $2$ batches. We attribute this to the larger per-batch sample support available with fewer batches, which partially offsets the induced shift; the pattern holds consistently across datasets.

\noindent\textbf{A note on reported percentages.} Throughout this section, percentage improvements (e.g.\ the $\Delta_3$, $\Delta_4$ columns of Tables~\ref{tab: ccca}, \ref{tab: c3fold}) are computed as $(\mu_{\text{PIcsC}} - \mu_{\text{baseline}})/\mu_{\text{baseline}} \times 100$, i.e.\ a \emph{relative} improvement in accuracy over the fragmented st-CV baseline for the matching setting (batch count or fold count). This is distinct from the comparison against prior state-of-the-art methods in Table~\ref{tab: meanacc} (Section~\ref{sota}), which reports point differences between $PIcsC$ and each baseline on the \emph{unfragmented} training set; the two tables answer different questions (does $PIcsC$ correct partition-induced shift, versus how does $PIcsC$ compare to prior work absent fragmentation) and are not directly comparable to one another.

\noindent\textbf{PIcs mitigation.} $PIcsC$ improves upon BL2 and mitigates PIcs on the no-covariate-shift datasets: Table~\ref{tab: ccca}, column $\Delta_3$, shows an increase in average accuracy of more than $10\%$ under the $20$-, $10$-, and $2$-batch fragmentation settings, sustained across datasets. We attribute this to $PIcsC$'s ability to retain information accumulated over the batch sequence while regularizing the model, which counteracts the loss of statistical support in each individual batch.

\begin{table}[htbp]
\centering
\caption{st-CV batch-wise accuracy}
\label{tab: bst-CV-full}
\renewcommand{\arraystretch}{1.0}
\setlength{\tabcolsep}{4pt}
\begin{tabular}{l c ccccc c c}
\toprule
\multirow{2}{*}{\textbf{Dataset}} & \multirow{2}{*}{\textbf{st-CV}} & \multicolumn{5}{c}{\textbf{Batchwise accuracy}} & \multirow{2}{*}{\textbf{Mean}} & \multirow{2}{*}{\textbf{var}} \\
\cmidrule(lr){3-7}
 & & \textbf{$B_1$} & \textbf{$B_2$} & \textbf{$B_{n/2}$} & \textbf{$B_{n-1}$} & \textbf{$B_n$} & $\mu_1$ & $\sigma_1^2$ \\
\midrule
\multicolumn{9}{c}{\textit{Training data = 5\%, Number of Batches = 20}} \\
\midrule
MNIST      & 94.8 & 89.3 & 87.9 & 89.9 & 88.9 & 88.8 & 88.7 & 0.49  \\
F-MNIST    & 83.1 & 73.7 & 74.2 & 72.5 & 74.0 & 70.6 & 72.9 & 1.94  \\
CIFAR-10   & 71.5 & 49.0 & 50.3 & 50.7 & 51.2 & 54.5 & 49.9 & 9.67  \\
CIFAR-100  & 38.2 & 16.5 & 18.1 & 19.3 & 20.4 & 22.7 & 18.3 & 9.17  \\
\cmidrule(lr){1-9}
P-MNIST    & 95.1 & 86.1 & 88.9 & 88.7 & 87.2 & 88.3 & 88.4 & 1.41  \\
K-MNIST    & 75.4 & 63.4 & 62.2 & 66.7 & 58.3 & 63.4 & 63.4 & 5.59  \\
CIFAR10-C  & 63.9 & 20.1 & 16.3 & 16.1 & 14.9 & 10.2 & 16.2 & 10.4  \\
CIFAR100-C & 28.8 & 16.3 & 19.1 & 19.7 & 21.6 & 22.3 & 18.4 & 14.2  \\
\midrule
\multicolumn{9}{c}{\textit{Training data = 10\%, Number of Batches = 10}} \\
\midrule
MNIST      & 94.8 & 91.7 & 91.1 & 90.2 & 89.3 & 91.5 & 91.1 & 1.06  \\
F-MNIST    & 83.1 & 69.9 & 75.2 & 73.4 & 74.4 & 71.7 & 73.7 & 9.22  \\
CIFAR-10   & 71.5 & 52.8 & 52.9 & 53.7 & 54.5 & 54.1 & 52.6 & 4.87  \\
CIFAR-100  & 38.2 & 20.3 & 22.2 & 23.3 & 24.7 & 21.2 & 21.5 & 5.51  \\
\cmidrule(lr){1-9}
P-MNIST    & 95.1 & 92.8 & 91.8 & 91.2 & 91.4 & 91.3 & 91.5 & 0.62  \\
K-MNIST    & 75.4 & 63.6 & 64.1 & 64.9 & 65.2 & 64.5 & 64.0 & 1.15  \\
CIFAR10-C  & 63.9 & 17.6 & 18.4 & 12.2 & 17.4 & 12.9 & 22.6 & 16.8  \\
CIFAR100-C & 28.8 & 21.9 & 25.2 & 26.6 & 27.1 & 20.5 & 22.8 & 16.3  \\
\midrule
\multicolumn{9}{c}{\textit{Training data = 20\%, Number of Batches = 5}} \\
\midrule
MNIST      & 94.8 & 92.2 & 93.4 & 91.2 & 90.2 & $\textendash$ & 91.9 & 1.59  \\
F-MNIST    & 83.1 & 74.1 & 72.0 & 75.3 & 77.2 & $\textendash$ & 75.1 & 4.40  \\
CIFAR-10   & 71.5 & 38.5 & 37.5 & 36.8 & 37.9 & $\textendash$ & 37.5 & 0.50  \\
CIFAR-100  & 38.2 & 22.3 & 23.5 & 24.4 & 24.9 & $\textendash$ & 22.9 & 3.43  \\
\cmidrule(lr){1-9}
P-MNIST    & 95.1 & 91.7 & 91.4 & 91.2 & 93.5 & $\textendash$ & 91.8 & 1.01  \\
K-MNIST    & 75.4 & 69.2 & 67.7 & 65.5 & 66.8 & $\textendash$ & 67.5 & 2.02  \\
CIFAR10-C  & 63.9 & 62.0 & 61.8 & 60.7 & 67.5 & $\textendash$ & 62.0 & 9.43  \\
CIFAR100-C & 28.8 & 23.2 & 23.2 & 26.7 & 26.1 & $\textendash$ & 23.8 & 5.77  \\
\midrule
\multicolumn{9}{c}{\textit{Training data = 25\%, Number of Batches = 4}} \\
\midrule
MNIST      & 94.8 & 93.6 & 92.5 & 91.3 & $\textendash$ & $\textendash$ & 92.4 & 0.92  \\
F-MNIST    & 83.1 & 77.5 & 78.1 & 78.5 & $\textendash$ & $\textendash$ & 77.2 & 3.12  \\
CIFAR-10   & 71.5 & 52.7 & 56.9 & 58.1 & $\textendash$ & $\textendash$ & 54.9 & 7.02  \\
CIFAR-100  & 38.2 & 22.4 & 23.9 & 24.8 & $\textendash$ & $\textendash$ & 22.8 & 2.70  \\
\cmidrule(lr){1-9}
P-MNIST    & 95.1 & 92.2 & 90.5 & 91.6 & $\textendash$ & $\textendash$ & 91.6 & 0.61  \\
K-MNIST    & 75.4 & 69.2 & 69.5 & 68.8 & $\textendash$ & $\textendash$ & 69.5 & 0.60  \\
CIFAR10-C  & 63.9 & 62.1 & 63.1 & 62.6 & $\textendash$ & $\textendash$ & 61.4 & 4.81  \\
CIFAR100-C & 28.8 & 23.4 & 23.8 & 24.5 & $\textendash$ & $\textendash$ & 22.8 & 3.64  \\
\midrule
\multicolumn{9}{c}{\textit{Training data = 50\%, Number of Batches = 2}} \\
\midrule
MNIST      & 94.8 & 93.7 & 93.3 & $\textendash$ & $\textendash$ & $\textendash$ & 93.5 & 0.08  \\
F-MNIST    & 83.1 & 79.7 & 80.0 & $\textendash$ & $\textendash$ & $\textendash$ & 79.8 & 0.04  \\
CIFAR-10   & 71.5 & 60.1 & 56.2 & $\textendash$ & $\textendash$ & $\textendash$ & 58.1 & 3.81  \\
CIFAR-100  & 38.2 & 25.4 & 23.2 & $\textendash$ & $\textendash$ & $\textendash$ & 24.3 & 1.21  \\
\cmidrule(lr){1-9}
P-MNIST    & 95.1 & 93.7 & 93.3 & $\textendash$ & $\textendash$ & $\textendash$ & 93.5 & 0.08  \\
K-MNIST    & 75.4 & 73.3 & 73.6 & $\textendash$ & $\textendash$ & $\textendash$ & 73.5 & 0.04  \\
CIFAR10-C  & 63.9 & 65.5 & 63.2 & $\textendash$ & $\textendash$ & $\textendash$ & 64.3 & 1.32  \\
CIFAR100-C & 28.8 & 27.5 & 25.3 & $\textendash$ & $\textendash$ & $\textendash$ & 26.4 & 1.21  \\
\bottomrule
\end{tabular}
\end{table}

\begin{table}[htbp]
\centering
\caption{$PIcsC$ Batchwise}
\label{tab: ccca}
\renewcommand{\arraystretch}{1.0}
\setlength{\tabcolsep}{4pt}
\begin{tabular}{l c cccccc c c c}
\toprule
\multirow{2}{*}{\textbf{Dataset}} & \multirow{2}{*}{\textbf{$PIcsC$}} & \multicolumn{6}{c}{\textbf{Batchwise accuracy}} & \multirow{2}{*}{\textbf{Mean}} & \multirow{2}{*}{\textbf{var}} & \multirow{2}{*}{\boldmath$\Delta_3$} \\
\cmidrule(lr){3-8}
 & & \textbf{$B_1$} & \textbf{$B_2$} & \textbf{$B_3$} & \textbf{$B_{n/2}$} & \textbf{$B_{n-1}$} & \textbf{$B_n$} & & & \\
\midrule
\multicolumn{11}{c}{\textit{Training data = 5\%, Number of Batches = 20}} \\
\midrule
MNIST      & 97.9 & 90.7 & 90.6 & 91.0 & 91.7 & 91.4 & 91.8 & 91.2 & 0.09  & $\uparrow$2.81  \\
F-MNIST    & 88.4 & 81.5 & 81.7 & 81.2 & 81.4 & 81.5 & 81.9 & 81.5 & 0.06  & $\uparrow$11.7  \\
CIFAR-10   & 87.7 & 50.9 & 51.4 & 52.2 & 48.9 & 50.3 & 57.4 & 51.8 & 7.18  & $\uparrow$3.81  \\
CIFAR-100  & 58.7 & 23.9 & 18.2 & 18.5 & 17.8 & 23.9 & 18.3 & 20.1 & 7.26  & $\uparrow$9.83  \\
\cmidrule(lr){1-11}
P-MNIST    & 97.6 & 91.1 & 89.8 & 90.2 & 90.4 & 91.5 & 90.7 & 90.3 & 0.26  & $\uparrow$2.14  \\
K-MNIST    & 89.2 & 68.5 & 69.4 & 67.2 & 68.0 & 67.8 & 66.9 & 68.4 & 0.63  & $\uparrow$7.88  \\
CIFAR10-C  & 73.3 & 46.4 & 54.3 & 57.8 & 61.1 & 61.5 & 61.8 & 57.2 & 30.1  & $\uparrow$253   \\
CIFAR100-C & 39.4 & 11.9 & 17.2 & 18.5 & 21.3 & 22.1 & 24.9 & 19.3 & 17.1  & $\uparrow$4.89  \\
\midrule
\multicolumn{11}{c}{\textit{Training data = 10\%, Number of Batches = 10}} \\
\midrule
MNIST      & 97.9 & 91.9 & 91.7 & 91.2 & 91.8 & 91.3 & 91.8 & 91.7 & 0.08  & $\uparrow$0.65  \\
F-MNIST    & 88.4 & 79.5 & 82.4 & 81.6 & 79.5 & 82.3 & 81.9 & 81.2 & 1.21  & $\uparrow$10.1  \\
CIFAR-10   & 87.7 & 52.1 & 53.1 & 48.5 & 59.3 & 52.5 & 55.7 & 53.5 & 11.1  & $\uparrow$1.71  \\
CIFAR-100  & 58.7 & 27.2 & 25.8 & 20.4 & 17.0 & 21.9 & 22.8 & 22.5 & 11.3  & $\uparrow$4.65  \\
\cmidrule(lr){1-11}
P-MNIST    & 97.6 & 91.6 & 91.9 & 91.3 & 91.6 & 90.1 & 91.2 & 91.5 & 0.31  & 0.00            \\
K-MNIST    & 89.2 & 71.4 & 70.4 & 71.7 & 70.7 & 70.5 & 70.9 & 70.9 & 0.71  & $\uparrow$10.7  \\
CIFAR10-C  & 73.3 & 52.7 & 59.9 & 61.9 & 64.4 & 66.1 & 65.7 & 61.7 & 21.1  & $\uparrow$173   \\
CIFAR100-C & 39.4 & 16.2 & 22.1 & 24.8 & 27.2 & 26.8 & 27.3 & 21.1 & 15.6  & $\downarrow$7.45\\
\midrule
\multicolumn{11}{c}{\textit{Training data = 20\%, Number of Batches = 5}} \\
\midrule
MNIST      & 97.9 & 93.6 & 93.8 & 94.3 & 93.7 & 93.8 & $\textendash$ & 93.8 & 0.07  & $\uparrow$2.06  \\
F-MNIST    & 88.4 & 82.8 & 83.1 & 82.1 & 81.4 & 82.6 & $\textendash$ & 82.4 & 0.44  & $\uparrow$6.73  \\
CIFAR-10   & 87.7 & 50.3 & 56.2 & 53.5 & 57.8 & 59.9 & $\textendash$ & 55.5 & 11.3  & $\uparrow$48.0  \\
CIFAR-100  & 58.7 & 34.2 & 35.4 & 33.9 & 34.9 & 34.7 & $\textendash$ & 34.6 & 11.7  & $\uparrow$51.1  \\
\cmidrule(lr){1-11}
P-MNIST    & 97.6 & 94.1 & 93.9 & 93.6 & 94.1 & 94.3 & $\textendash$ & 94.0 & 0.07  & $\uparrow$0.53  \\
K-MNIST    & 89.2 & 75.4 & 76.3 & 75.8 & 75.1 & 75.6 & $\textendash$ & 75.6 & 0.21  & $\uparrow$12.0  \\
CIFAR10-C  & 73.3 & 58.4 & 63.3 & 64.3 & 66.5 & 66.2 & $\textendash$ & 63.7 & 8.53  & $\downarrow$3.80\\
CIFAR100-C & 39.4 & 22.1 & 25.4 & 27.4 & 28.9 & 29.7 & $\textendash$ & 26.7 & 7.43  & $\uparrow$12.1  \\
\midrule
\multicolumn{11}{c}{\textit{Training data = 25\%, Number of Batches = 4}} \\
\midrule
MNIST      & 97.9 & 94.4 & 94.4 & 94.3 & 94.4 & $\textendash$ & $\textendash$ & 94.4 & 0.003 & $\uparrow$2.16  \\
F-MNIST    & 88.4 & 82.8 & 83.2 & 83.6 & 83.5 & $\textendash$ & $\textendash$ & 83.3 & 0.13  & $\uparrow$4.38  \\
CIFAR-10   & 87.7 & 56.8 & 57.3 & 62.2 & 63.4 & $\textendash$ & $\textendash$ & 59.9 & 8.47  & $\uparrow$9.10  \\
CIFAR-100  & 58.7 & 33.9 & 34.1 & 34.7 & 33.5 & $\textendash$ & $\textendash$ & 34.1 & 0.18  & $\uparrow$49.5  \\
\cmidrule(lr){1-11}
P-MNIST    & 97.6 & 94.4 & 94.3 & 94.5 & 94.5 & $\textendash$ & $\textendash$ & 94.4 & 0.009 & $\uparrow$3.05  \\
K-MNIST    & 89.2 & 77.2 & 75.5 & 77.6 & 75.3 & $\textendash$ & $\textendash$ & 76.4 & 1.37  & $\uparrow$11.0  \\
CIFAR10-C  & 73.3 & 60.9 & 64.4 & 66.8 & 68.3 & $\textendash$ & $\textendash$ & 65.1 & 7.81  & $\uparrow$6.02  \\
CIFAR100-C & 39.4 & 24.7 & 28.2 & 29.7 & 31.9 & $\textendash$ & $\textendash$ & 28.6 & 6.86  & $\uparrow$25.4  \\
\midrule
\multicolumn{11}{c}{\textit{Training data = 50\%, Number of Batches = 2}} \\
\midrule
MNIST      & 97.9 & 95.9 & 96.1 & $\textendash$ & $\textendash$ & $\textendash$ & $\textendash$ & 96.0  & 0.02  & $\uparrow$2.67  \\
F-MNIST    & 88.4 & 84.2 & 84.4 & $\textendash$ & $\textendash$ & $\textendash$ & $\textendash$ & 84.3  & 0.02  & $\uparrow$5.63  \\
CIFAR-10   & 87.7 & 76.3 & 80.6 & $\textendash$ & $\textendash$ & $\textendash$ & $\textendash$ & 78.4  & 4.62  & $\uparrow$34.9  \\
CIFAR-100  & 58.7 & 39.8 & 39.9 & $\textendash$ & $\textendash$ & $\textendash$ & $\textendash$ & 39.85 & 0.002 & $\uparrow$63.9  \\
\cmidrule(lr){1-11}
P-MNIST    & 97.6 & 95.7 & 96.1 & $\textendash$ & $\textendash$ & $\textendash$ & $\textendash$ & 95.9  & 0.08  & $\uparrow$2.56  \\
K-MNIST    & 89.2 & 79.3 & 80.4 & $\textendash$ & $\textendash$ & $\textendash$ & $\textendash$ & 79.8  & 0.61  & $\uparrow$8.57  \\
CIFAR10-C  & 73.3 & 65.5 & 68.6 & $\textendash$ & $\textendash$ & $\textendash$ & $\textendash$ & 67.1  & 2.40  & $\uparrow$4.35  \\
CIFAR100-C & 39.4 & 31.2 & 34.8 & $\textendash$ & $\textendash$ & $\textendash$ & $\textendash$ & 33.0  & 3.24  & $\uparrow$25.0  \\
\bottomrule
\end{tabular}
\end{table}

\noindent\textbf{Double mitigation.} Under the more challenging condition where natural shift is already present and is further aggravated by fragmentation, $PIcsC$ still improves upon BL2: Table~\ref{tab: ccca} shows a performance increase of more than $25\%$ under the $20$-, $10$-, and $2$-batch settings. We refer to this compound effect correcting both natural and partition-induced shift simultaneously as \textit{double mitigation}.

\noindent\textbf{Consistency.} Comparing a given batch position with and without remediation shows a consistent improvement across fragmentation settings and datasets. For example, under $20$-batch fragmentation on F-MNIST, batch $B_n$ reports $70.6\%$ accuracy without remediation (Table~\ref{tab: bst-CV-full}) and $81.9\%$ with remediation (Table~\ref{tab: ccca})  an improvement of $16\%$  consistent with $PIcsC$ regularizing model parameters from the current fragment while sustaining performance across the sequence.

\subsection{E3/E4: Fold-wise aggravation and correction of PIcs}
\label{sec:e3e4-results}

\textbf{E3 -- aggravation.} Table~\ref{tab: cccakeel} (tabular, KEEL) and Table~\ref{tab:st-CVfold} (image) report st-CV accuracy in the fold setting, against the BL1 baseline. Accuracy falls as a dataset is fragmented into more folds, consistent with covariate shift induced by the fragmentation itself; we report average accuracy per fold setting and per individual fold, denoted $\mu_3,\mu_4,\mu_5$ for st-CV with $2$, $5$, and $10$ folds respectively.

\noindent\textbf{E4 -- correction.} Tables~\ref{tab: c3keel} and \ref{tab: c3fold} report $PIcsC$ accuracy in the corresponding fold settings. On tabular datasets (Table~\ref{tab: c3keel}), $\Delta_5 = \mu_3-\mu_6$, $\Delta_6=\mu_4-\mu_7$, and $\Delta_7=\mu_5-\mu_8$ report the difference in average accuracy between st-CV and $PIcsC$ at $k=2,5,10$ folds; $PIcsC$ improves accuracy by up to $28.3\%$ across these settings. On image datasets (Table~\ref{tab: c3fold}), $\Delta_4 = \mu_9-\mu_{10}$ reports the corresponding difference; $PIcsC$ improves accuracy by up to $43.2\%$. We treat st-CV foldwise as the secondary baseline for these comparisons.

\subsection{E5: Comparison with the state of the art}
\label{sota}
Table~\ref{tab: meanacc} compares $PIcsC$ against IW, IWst-CV, KMM, and DIW (Section~\ref{bnchmrks}). $PIcsC$ improves over the strongest competitor, DIW (column $\Delta_2$), by up to $9.5\%$ on no-covariate-shift datasets and up to $20\%$ on natural-covariate-shift datasets, a pattern that holds consistently across datasets. We attribute DIW's shortfall to its iterative dependency between the feature extractor and classifier, a constraint $PIcsC$ does not have, since $PIcsC$ requires only the current fragment to update while retaining previously accumulated information. KMM's and IWst-CV's shortfalls are consistent with, respectively, KMM's assumption of an unbiased density ratio under covariate shift, and IWst-CV's known degradation in high dimensions (Section~\ref{bnchmrks}). The larger gains on natural-covariate-shift datasets are consistent with $PIcsC$'s double-mitigation property (Section~\ref{sec:e1e2-results}).

\begin{table}[htbp]
\centering
\caption{$PIcsC$ vs. state of the art (Section~\ref{sota}).}
\label{tab: meanacc}
\renewcommand{\arraystretch}{1.0}
\begin{tabular}{lcccccccc} 
\hline
\multirow{2}{*}{\textbf{Dataset}} & \textbf{Baseline} & \multicolumn{4}{c}{\textbf{SOTA}}  & \textbf{Ours}  & \multicolumn{2}{c}{\begin{tabular}[c]{@{}c@{}}$\Delta_1$ = ${PIcsC}$ - st-CV\\$\Delta_2$ = $PIcsC$ - DIW\end{tabular}}  \\
\cmidrule{2-9}
                         & st-CV       & IW   & IWst-CV & KMM  & DIW  & $PIcsC$ & $\Delta_1(\%)$  & $\Delta_2(\%)$                                                                               \\ 
\hline
MNIST                    & 94.8     & 85.1 & 76.9 & 11.8 & 98.0 & 97.9  & $\uparrow$ 3.16  & $\downarrow$ 0.11                                                                             \\
F-MNIST            & 82.3     & 80.2 & 72.3 & 10.3 & 87.2 & 88.4  & $\uparrow$ 6.90  & $\uparrow$ 1.37                                                                               \\
CIFAR10                 & 71.5     & 79.8 & 69.9 & 9.61 & 80.4 & 87.7  & $\uparrow$ 18.4 & $\uparrow$ 9.07                                                                               \\
CIFAR100                & 38.2     & 44.6 & 51.7 & 8.53 & 53.6 & 58.7  & $\uparrow$ 34.9 & $\uparrow$ 9.51                                                                               \\
\cmidrule{1-1}
P-MNIST           & 95.1     & 67.8 & 75.4 & 11.2 & 84.7 & 97.6  & $\uparrow$ 2.56  & $\uparrow$ 3.06                                                                              \\
K-MNIST          & 77.1     & 78.3 & 74.2 & 10.2 & 86.7 & 89.2  & $\uparrow$ 13.5 & $\uparrow$ 2.88                                                                               \\
CIFAR10-C                & 63.9     & 69.1 & 60.1 & 7.87 & 69.4 & 73.3  & $\uparrow$ 12.8  & $\uparrow$ 5.61                                                                               \\
CIFAR100-C               & 28.8     & 18.7 & 16.9 & 5.37 & 32.6 & 39.4  & $\uparrow$ 26.9 & $\uparrow$ 20.8                                                                               \\
\hline
\end{tabular}
\end{table}

\subsection{E6: Federated learning performance}
\label{sec:fl-results}
Table~\ref{tab: fedresults} reports results on the seven federated benchmarks of Section~\ref{sec:fed-setup}. The conditional mechanism (Section~\ref{sec:conditional}) outperforms FedAvg, FedProx, and SCAFFOLD on every dataset, by 3-5 percentage points, without any client-specific personalization component. Gains persist under the more heavily skewed client distributions (FEMNIST, SVHN), consistent with the same Fisher-KL trigger and Cram\'er-Rao regularizer that remediate PIcs across batches/folds (Sections~\ref{sec:e1e2-results}-\ref{sota}) transferring to client heterogeneity without any modification to Eqs.~\eqref{eq:tau}-\eqref{eq:loss-conditional}.

\begin{table}[htbp]
\centering
\caption{Federated learning performance on non-IID benchmarks. $PIcsC$ (conditional mechanism, Section~\ref{sec:conditional}) consistently outperforms classical federated baselines without client-specific personalization components.}
\label{tab: fedresults}
\renewcommand{\arraystretch}{1.0}
\begin{tabular}{lcccc}
\toprule
\textbf{Dataset} & \textbf{FedAvg} & \textbf{FedProx} & \textbf{SCAFFOLD} & \textbf{$PIcsC$}  \\
\midrule
FEMNIST          & 58.2 $\pm$ 3.1      & 60.1 $\pm$ 2.8       & 62.5 $\pm$ 2.5         & \textbf{64.8 $\pm$ 1.7}  \\
CIFAR-10         & 42.7 $\pm$ 4.5      & 45.3 $\pm$ 3.7       & 48.1 $\pm$ 3.2         & \textbf{49.5 $\pm$ 2.0}  \\
CIFAR-100        & 23.4 $\pm$ 2.8      & 25.1 $\pm$ 2.4       & 26.8 $\pm$ 2.1         & \textbf{28.3 $\pm$ 1.5}  \\
SVHN             & 61.5 $\pm$ 3.2      & 63.0 $\pm$ 2.9       & 65.2 $\pm$ 2.4         & \textbf{66.0 $\pm$ 1.9}  \\
EMNIST Digits    & 84.6 $\pm$ 1.7      & 86.0 $\pm$ 1.4       & 87.5 $\pm$ 1.2         & \textbf{88.6 $\pm$ 1.0}  \\
Amazon Reviews   & 76.3 $\pm$ 2.4      & 78.1 $\pm$ 2.1       & 80.4 $\pm$ 1.8         & \textbf{81.9 $\pm$ 1.4}  \\
Shakespeare      & 52.8 $\pm$ 3.0      & 54.6 $\pm$ 2.7       & 56.3 $\pm$ 2.3         & \textbf{57.5 $\pm$ 1.8}  \\
\bottomrule
\end{tabular}
\end{table}

\textbf{Federated setup details.} Datasets, client partitioning, and baseline configurations are as described in Sections~\ref{datasets}-\ref{sec:baselines}. Each client transmits only its diagonal FIM and current parameters to the aggregator; the server applies Algorithm~\ref{algo2}, updating the global FIM and regularizing (Eq.~\eqref{eq:loss-conditional}) only when the composite signal $\tau_t$ (Eq.~\eqref{eq:tau}) exceeds the calibrated threshold $\gamma$.
\label{sec:fed-setup}

\subsection{Calibrating the penalty}
\label{cp}
\begin{figure}[!htbp]
    \centering
    \includegraphics[width=0.85\textwidth]{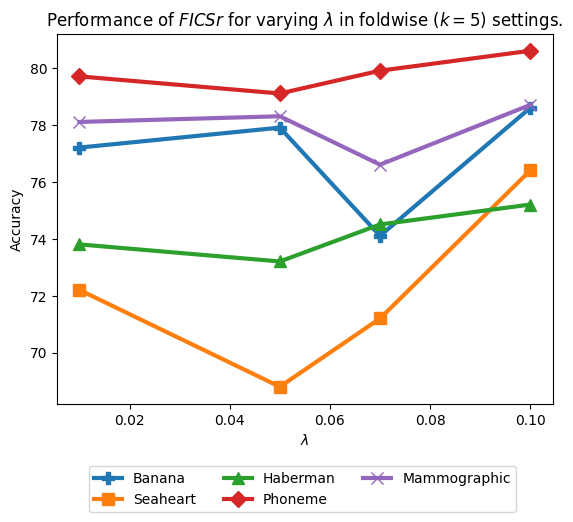}
    \caption{$PIcsC$ performance for calibrating $\lambda$ in fold-wise settings.}
    \label{fig:lambda}
\end{figure}

We calibrate the penalty strength $\lambda$ across batch/fold settings and report the result in Figure~\ref{fig:lambda}. $PIcsC$ performs best at $\lambda = 0.1$, which we use throughout the paper (Section~\ref{sec:impl}).

\begin{table}[htbp]
\centering
\caption{st-CV Foldwise Accuracy}
\label{tab:st-CVfold}
\renewcommand{\arraystretch}{1.0}
\setlength{\tabcolsep}{4pt}
\begin{tabular}{l cc ccccc c c}
\toprule
\multirow{2}{*}{\textbf{Dataset}} & \multicolumn{2}{c}{\textbf{Baseline}} & \multicolumn{5}{c}{\textbf{Foldwise accuracy}} & \multirow{2}{*}{\textbf{Mean}} & \multirow{2}{*}{\textbf{var}} \\
\cmidrule(lr){2-3} \cmidrule(lr){4-8}
 & \textbf{st-CV} & \textbf{$PIcsC$} & \textbf{$k_1$} & \textbf{$k_2$} & \textbf{$k_{n/2}$} & \textbf{$k_{n-1}$} & \textbf{$k_n$} & $\mu_9$ & $\sigma_3^2$ \\
\midrule
\multicolumn{10}{c}{\textit{Number of Folds = 10}} \\
\midrule
MNIST         & 94.8 & 97.9 & 95.1 & 94.8 & 94.1 & 95.3 & 94.6 & 94.9 & 0.17  \\
P-MNIST       & 95.1 & 97.6 & 94.1 & 94.5 & 94.3 & 94.3 & 93.5 & 94.0 & 0.11  \\
F-MNIST       & 83.1 & 88.4 & 83.9 & 84.2 & 82.6 & 84.2 & 83.7 & 83.7 & 0.35  \\
K-MNIST       & 75.4 & 89.2 & 85.5 & 85.6 & 86.2 & 86.6 & 86.2 & 85.9 & 0.17  \\
MNIST-C       & 97.7 & 99.2 & 94.1 & 95.1 & 96.2 & 95.8 & 96.9 & 95.9 & 0.92  \\
CIFAR-10      & 71.5 & 88.7 & 65.6 & 66.4 & 64.4 & 65.3 & 63.6 & 64.4 & 0.94  \\
CIFAR-100     & 38.2 & 58.7 & 31.4 & 29.9 & 30.2 & 32.2 & 30.7 & 31.8 & 0.69  \\
CIFAR10-C     & 63.9 & 73.3 & 66.6 & 67.5 & 65.5 & 66.1 & 67.7 & 66.4 & 0.69  \\
CIFAR100-C    & 28.8 & 39.4 & 33.4 & 32.8 & 32.8 & 32.1 & 31.9 & 32.6 & 0.29  \\
SVHN          & 85.1 & 90.8 & 87.3 & 84.6 & 19.2 & 84.4 & 86.2 & 79.1 & 707   \\
Caltech101    & 52.3 & 51.7 & 54.8 & 58.6 & 53.5 & 52.9 & 55.7 & 55.7 & 4.02  \\
Tiny-ImageNet & 42.5 & 59.2 & 41.1 & 39.3 & 33.9 & 40.8 & 38.2 & 38.6 & 6.76  \\
STL-10        & 43.6 & 56.2 & 45.2 & 42.7 & 40.1 & 39.9 & 42.5 & 42.5 & 3.79  \\
\midrule
\multicolumn{10}{c}{\textit{Number of Folds = 5}} \\
\midrule
MNIST         & 94.8 & 97.9 & 94.7 & 94.3 & 95.1 & 95.1 & 94.9 & 94.8 & 0.09  \\
P-MNIST       & 95.1 & 97.6 & 93.5 & 94.1 & 93.5 & 93.3 & 94.0 & 93.6 & 0.09  \\
F-MNIST       & 83.1 & 88.4 & 84.1 & 83.8 & 84.3 & 83.5 & 84.7 & 84.1 & 0.17  \\
K-MNIST       & 75.4 & 89.2 & 86.8 & 84.3 & 86.6 & 85.8 & 85.7 & 85.8 & 0.77  \\
MNIST-C       & 97.7 & 99.2 & 93.1 & 94.4 & 94.5 & 96.2 & 95.9 & 94.8 & 1.26  \\
CIFAR-10      & 71.5 & 88.7 & 59.7 & 63.4 & 61.3 & 61.6 & 62.7 & 61.8 & 1.61  \\
CIFAR-100     & 38.2 & 58.7 & 29.8 & 31.3 & 31.9 & 33.5 & 31.1 & 31.5 & 1.44  \\
CIFAR10-C     & 63.9 & 73.3 & 64.9 & 65.8 & 65.1 & 66.2 & 65.1 & 65.4 & 0.25  \\
CIFAR100-C    & 28.8 & 39.4 & 32.3 & 32.3 & 32.0 & 31.5 & 31.5 & 31.9 & 0.13  \\
SVHN          & 85.1 & 90.8 & 83.0 & 84.2 & 84.9 & 86.2 & 85.7 & 84.8 & 1.28  \\
Caltech101    & 52.3 & 51.7 & 55.9 & 52.3 & 56.4 & 50.7 & 56.2 & 54.3 & 5.51  \\
Tiny-ImageNet & 42.5 & 59.2 & 41.8 & 33.7 & 39.3 & 40.5 & 34.8 & 51.1 & 10.2  \\
STL-10        & 43.6 & 56.2 & 44.8 & 41.5 & 46.5 & 41.7 & 43.7 & 43.6 & 3.57  \\
\midrule
\multicolumn{10}{c}{\textit{Number of Folds = 2}} \\
\midrule
MNIST         & 94.8 & 97.9 & 91.4 & 92.4 & $\textendash$ & $\textendash$ & $\textendash$ & 93.7 & 0.25  \\
P-MNIST       & 95.1 & 97.6 & 92.5 & 92.4 & $\textendash$ & $\textendash$ & $\textendash$ & 92.4 & 0.003 \\
F-MNIST       & 83.1 & 88.4 & 83.9 & 84.2 & $\textendash$ & $\textendash$ & $\textendash$ & 84.1 & 0.02  \\
K-MNIST       & 75.4 & 89.2 & 82.5 & 84.4 & $\textendash$ & $\textendash$ & $\textendash$ & 83.4 & 0.91  \\
MNIST-C       & 97.7 & 99.2 & 93.1 & 94.1 & $\textendash$ & $\textendash$ & $\textendash$ & 93.6 & 0.25  \\
CIFAR-10      & 71.5 & 88.7 & 57.3 & 58.8 & $\textendash$ & $\textendash$ & $\textendash$ & 58.1 & 0.56  \\
CIFAR-100     & 38.2 & 58.7 & 26.3 & 25.7 & $\textendash$ & $\textendash$ & $\textendash$ & 26.1 & 0.09  \\
CIFAR10-C     & 63.9 & 73.3 & 63.1 & 64.9 & $\textendash$ & $\textendash$ & $\textendash$ & 64.2 & 0.81  \\
CIFAR100-C    & 28.8 & 39.4 & 28.2 & 28.1 & $\textendash$ & $\textendash$ & $\textendash$ & 28.1 & 0.003 \\
SVHN          & 85.1 & 90.8 & 82.1 & 19.0 & $\textendash$ & $\textendash$ & $\textendash$ & 50.5 & 995   \\
Caltech101    & 52.3 & 51.7 & 50.9 & 49.3 & $\textendash$ & $\textendash$ & $\textendash$ & 50.1 & 0.64  \\
Tiny-ImageNet & 42.5 & 59.2 & 39.6 & 32.8 & $\textendash$ & $\textendash$ & $\textendash$ & 36.2 & 11.5  \\
STL-10        & 43.6 & 56.2 & 33.8 & 39.9 & $\textendash$ & $\textendash$ & $\textendash$ & 36.8 & 18.6  \\
\bottomrule
\end{tabular}
\end{table}

\begin{table}[htbp]
\centering
\caption{$PIcsC$ Foldwise Accuracy}
\label{tab: c3fold}
\renewcommand{\arraystretch}{1.0}
\setlength{\tabcolsep}{4pt}
\begin{tabular}{l cc ccccc c c c}
\toprule
\multirow{2}{*}{\textbf{Dataset}} & \multicolumn{2}{c}{\textbf{Baseline}} & \multicolumn{5}{c}{\textbf{Foldwise accuracy}} & \multirow{2}{*}{\textbf{Mean}} & \multirow{2}{*}{\textbf{var}} & \multirow{2}{*}{\boldmath$\Delta_4$} \\
\cmidrule(lr){2-3} \cmidrule(lr){4-8}
 & \textbf{st-CV} & \textbf{$PIcsC$} & \textbf{$k_1$} & \textbf{$k_2$} & \textbf{$k_{n/2}$} & \textbf{$k_{n-1}$} & \textbf{$k_n$} & $\mu_{10}$ & $\sigma_4^2$ & $\Delta_4(\%)$ \\
\midrule
\multicolumn{11}{c}{\textit{Number of Folds = 10}} \\
\midrule
MNIST        & 94.8 & 97.9 & 97.8 & 97.3 & 97.4 & 97.2 & 97.4 & 97.4  & 0.04  & $\uparrow$2.50  \\
P-MNIST      & 95.1 & 97.6 & 91.2 & 91.1 & 91.3 & 91.2 & 90.9 & 91.1  & 0.02  & $\downarrow$2.90\\
F-MNIST      & 83.1 & 88.4 & 88.9 & 88.5 & 89.1 & 89.5 & 88.5 & 88.9  & 0.14  & $\uparrow$5.20  \\
K-MNIST      & 75.4 & 89.2 & 94.8 & 94.7 & 94.8 & 95.2 & 94.8 & 94.8  & 0.03  & $\uparrow$8.90  \\
MNIST-C      & 97.7 & 99.2 & 97.8 & 97.7 & 97.5 & 97.7 & 97.6 & 97.6  & 0.01  & $\uparrow$1.70  \\
CIFAR-10     & 71.5 & 88.7 & 85.7 & 85.4 & 85.5 & 85.4 & 85.3 & 85.4  & 0.02  & $\uparrow$21.0  \\
CIFAR-100    & 38.2 & 58.7 & 63.3 & 63.8 & 64.3 & 62.9 & 39.2 & 58.7  & 95.2  & $\uparrow$26.9  \\
CIFAR10-C    & 63.9 & 73.3 & 71.1 & 69.9 & 71.3 & 70.5 & 69.2 & 70.4  & 0.60  & $\uparrow$4.00  \\
CIFAR100-C   & 28.8 & 39.4 & 32.7 & 22.8 & 38.5 & 33.0 & 35.1 & 32.4  & 27.5  & $\downarrow$0.20\\
SVHN         & 85.1 & 90.8 & 95.6 & 95.9 & 95.8 & 95.4 & 95.5 & 95.6  & 0.034 & $\uparrow$16.5  \\
Caltech101   & 52.3 & 51.7 & 95.6 & 95.3 & 95.9 & 95.7 & 60.4 & 88.5  & 198   & $\uparrow$32.8  \\
Tiny-ImageNet& 42.5 & 59.2 & 57.9 & 56.7 & 49.1 & 53.4 & 57.5 & 54.8  & 10.9  & $\uparrow$16.2  \\
STL-10       & 43.6 & 56.2 & 51.3 & 50.6 & 52.7 & 49.4 & 47.2 & 50.2  & 4.33  & $\uparrow$7.70  \\
\midrule
\multicolumn{11}{c}{\textit{Number of Folds = 5}} \\
\midrule
MNIST        & 94.8 & 97.9 & 97.5  & 97.3  & 97.1  & 97.3  & 97.2 & 97.2  & 0.017 & $\uparrow$2.40  \\
P-MNIST      & 95.1 & 97.6 & 97.5  & 97.5  & 97.6  & 97.6  & 97.5 & 97.54 & 0.002 & $\uparrow$3.94  \\
F-MNIST      & 83.1 & 88.4 & 88.3  & 88.5  & 88.6  & 88.7  & 88.8 & 88.5  & 0.029 & $\uparrow$4.40  \\
K-MNIST      & 75.4 & 89.2 & 94.6  & 94.7  & 94.7  & 94.7  & 95.1 & 94.7  & 0.03  & $\uparrow$8.90  \\
MNIST-C      & 97.7 & 99.2 & 97.2  & 97.3  & 97.3  & 97.4  & 97.4 & 97.3  & 0.005 & $\uparrow$2.50  \\
CIFAR-10     & 71.5 & 88.7 & 71.43 & 71.65 & 72.18 & 72.27 & 73.47& 72.2  & 0.502 & $\uparrow$10.4  \\
CIFAR-100    & 38.2 & 58.7 & 67.8  & 68.1  & 67.7  & 67.9  & 39.2 & 62.1  & 132   & $\uparrow$30.6  \\
CIFAR10-C    & 63.9 & 73.3 & 91.2  & 91.5  & 91.2  & 91.4  & 69.6 & 86.9  & 75.5  & $\uparrow$21.5  \\
CIFAR100-C   & 28.8 & 39.4 & 27.5  & 38.1  & 28.8  & 31.4  & 33.0 & 31.7  & 13.7  & $\downarrow$0.20\\
SVHN         & 85.1 & 90.8 & 96.9  & 97.1  & 96.8  & 96.7  & 92.9 & 96.1  & 2.54  & $\uparrow$11.3  \\
Caltech101   & 52.3 & 51.7 & 94.9  & 95.1  & 94.2  & 94.6  & 57.6 & 87.2  & 220   & $\uparrow$32.9  \\
Tiny-ImageNet& 42.5 & 59.2 & 56.3  & 51.7  & 53.2  & 52.8  & 38.3 & 50.5  & 39.3  & $\downarrow$0.60\\
STL-10       & 43.6 & 56.2 & 52.9  & 48.7  & 50.6  & 53.3  & 50.9 & 51.2  & 3.49  & $\uparrow$7.50  \\
\midrule
\multicolumn{11}{c}{\textit{Number of Folds = 2}} \\
\midrule
MNIST        & 94.8 & 97.9 & 96.6 & 96.7 & $\textendash$ & $\textendash$ & $\textendash$ & 96.6  & 0.002 & $\uparrow$2.90  \\
P-MNIST      & 95.1 & 97.6 & 96.9 & 96.9 & $\textendash$ & $\textendash$ & $\textendash$ & 96.9  & 0.00  & $\uparrow$4.50  \\
F-MNIST      & 83.1 & 88.4 & 87.1 & 87.0 & $\textendash$ & $\textendash$ & $\textendash$ & 87.05 & 0.002 & $\uparrow$2.95  \\
K-MNIST      & 75.4 & 89.2 & 93.7 & 93.6 & $\textendash$ & $\textendash$ & $\textendash$ & 93.6  & 0.002 & $\uparrow$10.2  \\
MNIST-C      & 97.7 & 99.2 & 96.7 & 96.7 & $\textendash$ & $\textendash$ & $\textendash$ & 96.7  & 0.00  & $\uparrow$3.10  \\
CIFAR-10     & 71.5 & 88.7 & 69.2 & 67.9 & $\textendash$ & $\textendash$ & $\textendash$ & 68.5  & 0.43  & $\uparrow$10.4  \\
CIFAR-100    & 38.2 & 58.7 & 55.2 & 32.6 & $\textendash$ & $\textendash$ & $\textendash$ & 43.9  & 128   & $\uparrow$17.8  \\
CIFAR10-C    & 63.9 & 73.3 & 91.3 & 66.5 & $\textendash$ & $\textendash$ & $\textendash$ & 78.9  & 154   & $\uparrow$14.7  \\
CIFAR100-C   & 28.8 & 39.4 & 39.3 & 22.8 & $\textendash$ & $\textendash$ & $\textendash$ & 31.1  & 68.1  & $\uparrow$3.00  \\
SVHN         & 85.1 & 90.8 & 96.0 & 91.5 & $\textendash$ & $\textendash$ & $\textendash$ & 93.7  & 5.06  & $\uparrow$43.2  \\
Caltech101   & 52.3 & 51.7 & 94.7 & 53.4 & $\textendash$ & $\textendash$ & $\textendash$ & 74.1  & 426   & $\uparrow$24.0  \\
Tiny-ImageNet& 42.5 & 59.2 & 55.2 & 45.7 & $\textendash$ & $\textendash$ & $\textendash$ & 50.4  & 22.5  & $\uparrow$14.2  \\
STL-10       & 43.6 & 56.2 & 43.1 & 44.8 & $\textendash$ & $\textendash$ & $\textendash$ & 43.9  & 1.44  & $\uparrow$6.10  \\
\bottomrule
\end{tabular}
\end{table}

\begin{sidewaystable}
\centering
\footnotesize
\renewcommand{\arraystretch}{1.15}
\setlength{\tabcolsep}{4pt}
\caption{st-CV foldwise}
\label{tab: cccakeel}
\begin{tabular}{l c cc c ccccc c ccccc c}
\toprule
\multirow{2}{*}{\textbf{Dataset}} & \multirow{2}{*}{\textbf{st-CV}} & \multicolumn{3}{c}{\boldmath$k=2$} & \multicolumn{6}{c}{\boldmath$k=5$} & \multicolumn{6}{c}{\boldmath$k=10$} \\
\cmidrule(lr){3-5} \cmidrule(lr){6-11} \cmidrule(lr){12-17}
 & & $k_1$ & $k_2$ & $\mu_3$ & $k_1$ & $k_2$ & $k_3$ & $k_4$ & $k_5$ & $\mu_4$ & $k_1$ & $k_2$ & $k_{n/2}$ & $k_{n-1}$ & $k_n$ & $\mu_5$ \\
\midrule
Appendicitis   & 98.1 & 97.6 & 97.6 & 97.6 & 97.8 & 98.8 & 96.7 & 97.8 & 98.5 & 97.9 & 96.2 & 99.2 & 97.1 & 97.8 & 98.5 & 98.0 \\
Australian     & 85.5 & 81.4 & 83.2 & 82.3 & 84.1 & 84.1 & 87.6 & 84.7 & 86.9 & 85.5 & 85.5 & 84.1 & 88.4 & 79.7 & 89.8 & 85.3 \\
Banana         & 77.9 & 70.9 & 75.8 & 73.4 & 71.4 & 70.4 & 76.6 & 72.5 & 75.8 & 73.3 & 70.7 & 70.0 & 72.1 & 73.0 & 70.3 & 71.7 \\
Bands          & 81.5 & 68.8 & 71.1 & 70.0 & 71.2 & 70.3 & 71.2 & 63.8 & 74.1 & 70.1 & 68.5 & 64.8 & 70.3 & 57.4 & 77.7 & 74.1 \\
Breast         & 65.5 & 59.4 & 57.3 & 58.3 & 62.1 & 61.4 & 57.8 & 56.1 & 66.6 & 60.8 & 65.5 & 58.6 & 65.5 & 62.1 & 64.2 & 61.5 \\
Bupa           & 54.5 & 64.2 & 51.8 & 58.1 & 63.6 & 54.5 & 81.8 & 36.3 & 63.6 & 60.0 & 33.3 & 33.3 & 80.0 & 60.0 & 40.0 & 61.3 \\
Chess          & 98.4 & 93.3 & 93.9 & 93.6 & 95.0 & 96.4 & 97.8 & 98.1 & 97.1 & 96.8 & 96.5 & 98.7 & 98.7 & 97.8 & 98.4 & 97.9 \\
CrX            & 84.7 & 75.3 & 73.1 & 74.2 & 79.7 & 79.7 & 73.9 & 81.8 & 84.1 & 79.8 & 79.7 & 79.7 & 76.8 & 84.1 & 81.1 & 82.4 \\
German Credit  & 70.5 & 69.8 & 68.6 & 69.2 & 75.5 & 70.0 & 65.0 & 73.5 & 72.5 & 71.3 & 72.0 & 77.0 & 65.0 & 78.0 & 70.0 & 71.7 \\
Haberman       & 69.4 & 73.8 & 70.5 & 72.2 & 70.9 & 77.1 & 73.7 & 75.4 & 72.1 & 73.8 & 77.4 & 77.4 & 58.1 & 73.3 & 76.6 & 74.1 \\
Statlog(Heart) & 83.3 & 59.2 & 51.8 & 55.5 & 64.8 & 64.8 & 64.8 & 57.4 & 62.9 & 62.9 & 66.6 & 62.9 & 66.6 & 51.8 & 66.6 & 66.6 \\
Hepatitis      & 74.2 & 73.1 & 70.1 & 71.6 & 80.6 & 70.9 & 74.2 & 74.2 & 77.4 & 75.4 & 68.7 & 68.7 & 80.0 & 66.6 & 80.0 & 76.7 \\
House-votes    & 90.8 & 88.5 & 86.2 & 87.4 & 90.8 & 89.6 & 91.9 & 85.1 & 83.9 & 88.2 & 90.9 & 95.4 & 95.3 & 90.6 & 79.1 & 88.1 \\
Ionosphere     & 84.5 & 61.4 & 70.8 & 66.1 & 80.3 & 64.2 & 80.0 & 80.0 & 84.2 & 77.7 & 82.8 & 57.1 & 71.4 & 91.4 & 82.8 & 78.9 \\
Mammographic   & 79.7 & 76.7 & 75.0 & 75.8 & 78.2 & 74.4 & 76.0 & 77.1 & 75.5 & 76.2 & 77.1 & 77.1 & 79.2 & 79.2 & 78.1 & 76.6 \\
Monk-2         & 94.6 & 65.4 & 70.8 & 68.1 & 65.1 & 75.6 & 59.4 & 83.7 & 71.1 & 71.1 & 69.6 & 78.5 & 57.1 & 89.1 & 72.7 & 72.5 \\
Mushroom       & 99.1 & 100.0& 96.7 & 98.3 & 100.0& 100.0& 100.0& 100.0& 100.0& 100.0& 100.0& 100.0& 100.0& 100.0& 100.0& 100.0\\
Phoneme        & 80.6 & 80.7 & 78.8 & 79.7 & 82.7 & 80.5 & 80.6 & 80.6 & 80.3 & 80.9 & 82.9 & 79.8 & 79.1 & 81.8 & 81.5 & 80.6 \\
Pima           & 74.0 & 71.4 & 69.5 & 70.4 & 72.1 & 76.6 & 73.3 & 78.4 & 72.5 & 74.6 & 76.6 & 84.4 & 68.8 & 76.6 & 71.1 & 74.9 \\
Saheart        & 73.1 & 75.7 & 65.8 & 70.7 & 77.4 & 70.9 & 80.4 & 68.4 & 60.8 & 71.6 & 78.7 & 69.5 & 78.2 & 65.2 & 65.2 & 70.5 \\
Sonar          & 73.8 & 80.7 & 75.0 & 77.8 & 88.1 & 85.7 & 80.9 & 82.9 & 73.1 & 82.2 & 80.9 & 85.7 & 85.7 & 76.1 & 65.0 & 81.6 \\
Spambase       & 94.1 & 92.3 & 93.3 & 92.8 & 93.5 & 93.2 & 91.7 & 93.2 & 93.4 & 93.1 & 93.6 & 93.2 & 95.0 & 94.5 & 93.6 & 93.2 \\
SPECT Heart    & 74.1 & 68.6 & 65.4 & 67.0 & 70.4 & 72.2 & 64.2 & 75.5 & 58.5 & 68.2 & 74.1 & 70.3 & 66.6 & 80.7 & 57.6 & 68.1 \\
Tic-Tac-Toe    & 73.9 & 59.1 & 64.1 & 61.5 & 63.5 & 60.4 & 72.9 & 63.4 & 61.7 & 64.4 & 68.7 & 65.6 & 73.9 & 55.2 & 65.2 & 64.7 \\
Titanic        & 73.4 & 78.0 & 77.0 & 77.5 & 72.5 & 78.8 & 78.8 & 77.2 & 77.7 & 77.1 & 77.7 & 78.6 & 75.9 & 76.3 & 76.8 & 76.8 \\
Wdbc           & 97.3 & 96.1 & 95.1 & 95.6 & 95.6 & 98.2 & 98.2 & 97.3 & 92.9 & 96.4 & 96.4 & 98.2 & 98.2 & 96.4 & 89.2 & 96.4 \\
Wisconsin      & 96.4 & 95.7 & 95.1 & 95.4 & 95.7 & 97.8 & 95.7 & 95.7 & 93.5 & 95.7 & 95.7 & 98.5 & 97.1 & 94.3 & 94.2 & 95.7 \\
\bottomrule
\end{tabular}
\end{sidewaystable}

\clearpage

\begin{sidewaystable}
\centering
\footnotesize
\renewcommand{\arraystretch}{1.15}
\setlength{\tabcolsep}{3pt}
\caption{$PIcsC$ foldwise}
\label{tab: c3keel}
\begin{tabular}{l c cc c ccccc c ccccc c ccc}
\toprule
\multirow{2}{*}{\textbf{Dataset}} & \multirow{2}{*}{\textbf{BL}} & \multicolumn{3}{c}{\boldmath$k=2$} & \multicolumn{6}{c}{\boldmath$k=5$} & \multicolumn{6}{c}{\boldmath$k=10$} & \multicolumn{3}{c}{\boldmath$\Delta$} \\
\cmidrule(lr){3-5} \cmidrule(lr){6-11} \cmidrule(lr){12-17} \cmidrule(lr){18-20}
 & & $k_1$ & $k_2$ & $\mu_6$ & $k_1$ & $k_2$ & $k_3$ & $k_4$ & $k_5$ & $\mu_7$ & $k_1$ & $k_2$ & $k_{n/2}$ & $k_{n-1}$ & $k_n$ & $\mu_8$ & $\Delta_5$ & $\Delta_6$ & $\Delta_7$ \\
\midrule
Appendicitis   & 99.6 & 99.1 & 98.8 & 98.9 & 98.5 & 100.0& 98.2 & 98.2 & 98.5 & 98.6 & 97.1 & 100.0& 98.5 & 100.0& 100.0& 99.2 & $\uparrow$1.30 & $\uparrow$0.70 & $\uparrow$1.20 \\
Australian     & 91.3 & 86.6 & 84.1 & 85.3 & 85.5 & 85.5 & 87.6 & 84.7 & 88.4 & 86.3 & 86.9 & 85.5 & 86.9 & 73.9 & 92.7 & 86.2 & $\uparrow$3.00 & $\uparrow$0.80 & $\uparrow$0.90 \\
Banana         & 76.3 & 73.1 & 82.1 & 77.6 & 75.2 & 70.9 & 77.5 & 72.1 & 76.6 & 74.5 & 82.6 & 56.0 & 76.6 & 70.5 & 81.5 & 74.1 & $\uparrow$4.20 & $\uparrow$1.20 & $\uparrow$2.40 \\
Bands          & 76.8 & 77.4 & 68.8 & 73.1 & 77.7 & 74.1 & 67.5 & 72.2 & 78.8 & 74.1 & 72.2 & 66.6 & 77.7 & 64.8 & 74.1 & 71.6 & $\uparrow$3.10 & $\uparrow$4.00 & $\downarrow$2.50 \\
Breast         & 63.7 & 58.7 & 60.1 & 59.4 & 46.5 & 61.4 & 59.6 & 52.6 & 64.9 & 57.1 & 51.7 & 58.6 & 58.6 & 46.4 & 60.7 & 59.4 & $\uparrow$1.10 & $\downarrow$3.70 & $\downarrow$2.10 \\
Bupa           & 72.7 & 57.1 & 59.2 & 58.2 & 72.7 & 63.6 & 54.5 & 36.3 & 63.6 & 58.2 & 66.6 & 33.3 & 0.00 & 20.0 & 40.0 & 55.0 & $\uparrow$0.00 & $\downarrow$1.80 & $\downarrow$6.30 \\
Chess          & 98.7 & 98.6 & 98.4 & 98.5 & 98.1 & 99.3 & 99.2 & 98.4 & 97.1 & 98.4 & 98.4 & 98.7 & 99.3 & 99.1 & 98.4 & 99.1 & $\uparrow$4.90 & $\uparrow$1.60 & $\uparrow$1.20 \\
CrX            & 84.1 & 84.3 & 87.2 & 85.7 & 86.9 & 86.9 & 84.7 & 90.5 & 83.3 & 86.5 & 86.9 & 91.3 & 82.6 & 86.9 & 86.9 & 86.9 & $\uparrow$11.5 & $\uparrow$6.70 & $\uparrow$4.50 \\
German Credit  & 74.0 & 72.2 & 71.6 & 71.9 & 75.5 & 71.0 & 68.0 & 72.5 & 75.0 & 72.4 & 73.0 & 76.0 & 66.0 & 74.0 & 76.0 & 72.8 & $\uparrow$2.70 & $\uparrow$1.10 & $\uparrow$1.10 \\
Haberman       & 72.5 & 77.1 & 70.5 & 73.8 & 69.4 & 75.4 & 73.7 & 77.1 & 75.4 & 74.2 & 67.7 & 80.6 & 64.5 & 73.3 & 73.3 & 74.1 & $\uparrow$1.60 & $\uparrow$0.40 & 0.00 \\
Statlog(Heart) & 87.1 & 77.1 & 80.7 & 78.8 & 85.2 & 72.2 & 88.8 & 85.2 & 75.9 & 81.5 & 92.5 & 74.1 & 85.1 & 85.1 & 77.7 & 82.9 & $\uparrow$23.3 & $\uparrow$18.6 & $\uparrow$16.3 \\
Hepatitis      & 87.9 & 87.2 & 67.5 & 77.3 & 74.2 & 80.6 & 58.1 & 83.8 & 70.9 & 73.5 & 87.5 & 68.7 & 93.3 & 80.0 & 80.0 & 82.7 & $\uparrow$5.70 & $\downarrow$1.90 & $\uparrow$6.00 \\
House-votes    & 94.2 & 92.2 & 94.9 & 93.5 & 91.9 & 94.2 & 94.2 & 93.1 & 96.5 & 94.1 & 88.6 & 95.4 & 95.3 & 93.1 & 97.6 & 94.2 & $\uparrow$6.10 & $\uparrow$5.90 & $\uparrow$6.10 \\
Ionosphere     & 85.9 & 82.3 & 85.7 & 84.1 & 85.9 & 88.5 & 85.7 & 85.7 & 85.7 & 86.3 & 82.8 & 88.5 & 80.0 & 88.5 & 85.7 & 86.6 & $\uparrow$18.0 & $\uparrow$8.60 & $\uparrow$7.70 \\
Mammographic   & 80.3 & 80.0 & 76.4 & 78.2 & 79.7 & 73.9 & 78.1 & 76.5 & 75.5 & 76.7 & 79.1 & 75.0 & 76.1 & 80.2 & 77.1 & 77.2 & $\uparrow$2.40 & $\uparrow$0.50 & $\uparrow$0.60 \\
Monk-2         & 99.1 & 92.8 & 100.0& 96.4 & 100.0& 90.1 & 92.7 & 100.0& 81.9 & 92.9 & 83.9 & 87.5 & 100.0& 89.1 & 85.4 & 90.5 & $\uparrow$28.3 & $\uparrow$21.8 & $\uparrow$18.0 \\
Mushroom       & 100.0& 99.9 & 99.7 & 99.8 & 97.8 & 99.1 & 100.0& 97.9 & 98.7 & 98.7 & 100.0& 100.0& 99.1 & 100.0& 98.7 & 99.4 & $\uparrow$1.50 & $\downarrow$1.30 & $\downarrow$0.60 \\
Phoneme        & 81.3 & 78.4 & 80.4 & 79.4 & 81.1 & 80.2 & 81.1 & 77.8 & 78.5 & 79.9 & 83.7 & 77.6 & 78.2 & 79.8 & 79.6 & 79.1 & $\downarrow$0.30 & $\downarrow$1.00 & $\downarrow$1.50 \\
Pima           & 75.3 & 77.3 & 75.2 & 76.3 & 77.9 & 77.9 & 72.7 & 77.7 & 76.4 & 76.5 & 79.2 & 84.4 & 66.6 & 76.6 & 76.3 & 76.6 & $\uparrow$5.90 & $\uparrow$1.90 & $\uparrow$1.70 \\
Saheart        & 77.4 & 73.2 & 70.9 & 72.1 & 75.2 & 67.7 & 79.3 & 69.5 & 64.1 & 71.2 & 72.3 & 73.9 & 65.2 & 69.5 & 71.7 & 71.4 & $\uparrow$1.40 & $\downarrow$0.40 & $\uparrow$0.90 \\
Sonar          & 85.7 & 83.6 & 78.8 & 81.3 & 80.9 & 78.5 & 85.7 & 85.3 & 80.5 & 82.2 & 66.6 & 71.4 & 95.2 & 66.6 & 60.0 & 76.9 & $\uparrow$3.50 & 0.00 & $\downarrow$4.70 \\
Spambase       & 93.8 & 92.5 & 93.5 & 93.0 & 93.3 & 93.5 & 93.1 & 92.8 & 93.8 & 93.3 & 93.4 & 93.1 & 95.2 & 93.4 & 94.1 & 93.7 & $\uparrow$0.20 & $\uparrow$0.20 & $\uparrow$0.50 \\
SPECT Heart    & 85.1 & 79.1 & 73.6 & 76.3 & 83.3 & 83.3 & 83.1 & 79.2 & 71.6 & 80.1 & 85.1 & 74.1 & 74.1 & 84.6 & 65.3 & 75.6 & $\uparrow$9.30 & $\uparrow$11.9 & $\uparrow$7.50 \\
Tic-Tac-Toe    & 77.1 & 71.8 & 67.2 & 69.5 & 69.7 & 77.6 & 73.9 & 70.6 & 72.2 & 72.8 & 79.2 & 79.2 & 80.2 & 66.6 & 78.9 & 74.1 & $\uparrow$8.00 & $\uparrow$8.40 & $\uparrow$9.40 \\
Titanic        & 75.6 & 77.4 & 78.1 & 77.8 & 73.4 & 80.4 & 79.7 & 78.4 & 78.1 & 78.1 & 78.2 & 80.0 & 76.8 & 77.3 & 76.8 & 77.9 & $\uparrow$0.30 & $\uparrow$1.00 & $\uparrow$1.10 \\
Wdbc           & 99.1 & 97.8 & 97.1 & 97.5 & 98.2 & 99.1 & 98.2 & 98.2 & 96.4 & 98.1 & 98.2 & 98.2 & 94.7 & 98.2 & 96.4 & 97.7 & $\uparrow$1.90 & $\uparrow$1.70 & $\uparrow$1.30 \\
Wisconsin      & 97.2 & 96.5 & 96.5 & 96.5 & 97.1 & 97.8 & 97.1 & 97.1 & 94.2 & 96.7 & 97.1 & 98.5 & 95.7 & 95.7 & 95.6 & 96.1 & $\uparrow$1.10 & $\uparrow$1.00 & $\uparrow$0.40 \\
\bottomrule
\end{tabular}
\end{sidewaystable}

\FloatBarrier

\subsection{Ablation study}
\label{sec:ablation}
We examine two axes of robustness: (a) increasing the amount of covariate shift by injecting noise into the covariates, and (b) increasing the number of fragments so that each fragment covers a smaller support of the underlying distribution.

\begin{figure}[!ht]
    \centering
    \includegraphics[width=0.85\textwidth]{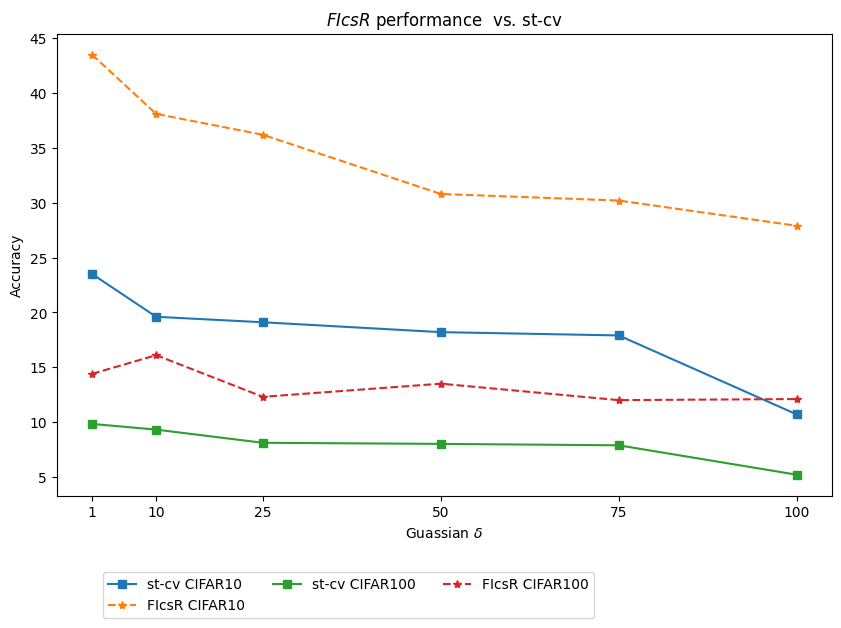}
    \caption{st-CV and $PIcsC$ accuracy under injected Gaussian covariate noise.}
    \label{fig:guassian}
\end{figure}

\textbf{Robustness to noise injection.} On CIFAR-10 and CIFAR-100 we inject Gaussian noise into the covariates using the pipeline of \cite{rabanser2019failing}, at standard deviations $\{1, 10, 25, 50, 75, 100\}$. Figure~\ref{fig:guassian} shows $PIcsC$ is comparatively robust across this range: at the highest noise level ($\sigma=100$) $PIcsC$ retains an accuracy advantage of up to $89\%$ over st-CV on CIFAR-10.

\begin{figure}[!ht]
    \centering
    \includegraphics[width=0.85\textwidth]{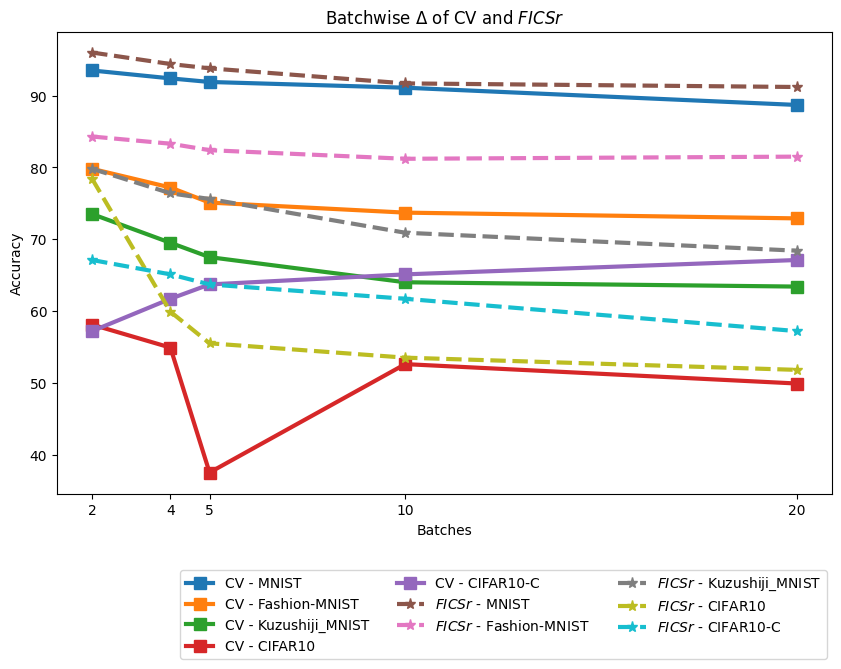}
    \caption{st-CV and $PIcsC$: $\Delta$ accuracy for varying numbers of batches.}
    \label{fig:delta}
    \end{figure}

\textbf{Robustness to fragment count.} Figure~\ref{fig:delta} compares $PIcsC$ and st-CV (\textbf{BL2}) as the number of batches increases. The accuracy gap $\Delta$ between $PIcsC$ and st-CV is smaller for $PIcsC$ across almost every fragmentation setting, with the exception of CIFAR10-C, indicating that $PIcsC$ degrades more gracefully than st-CV as the number of batches/folds grows.

\subsection{Discussion}
\label{sec:discussion}
Taken together, E1-E5 establish that fragmentation of a dataset for cross-validation  whether by batching or by $k$-fold splitting  induces a measurable covariate shift even on datasets with no shift prior to fragmentation, that this shift compounds with any natural shift already present, and that $PIcsC$'s Fisher-based penalty remediates both sources simultaneously (Sections~\ref{sec:e1e2-results}-\ref{sec:e3e4-results}) while remaining competitive with, and generally superior to, importance-weighting and kernel-based state-of-the-art baselines (Section~\ref{sota}). E6 extends the same estimator, unmodified except for the conditional trigger of Section~\ref{sec:conditional}, to the client/node instantiation, where it improves on established federated-optimization baselines that do not explicitly model inter-client distributional divergence (Section~\ref{sec:fl-results}). This supports our central claim (Section~\ref{intro}) that batch/fold and client/node fragmentation are two instantiations of a single underlying phenomenon, remediable by a single Fisher-KL estimator.\\

\noindent We note two limitations. First, the unconditional penalty (Algorithm~\ref{algo1}) is quadratic in the number of fragments and is therefore practical only when $k$ is small and bounded, as in our batch/fold experiments (Section~\ref{3}); the conditional mechanism (Algorithm~\ref{algo2}) addresses this for long or open-ended sequences but introduces two additional hyperparameters ($\alpha$, $\gamma$) that must be calibrated per deployment (Section~\ref{sec:impl}). Second, the Gaussian/CRLB approximation of Section~\ref{sec:fim-approx} is exact only under standard regularity conditions on the likelihood; our diagonal-FIM approximation further trades some estimation accuracy for the $\mathcal{O}(d)$ memory and communication cost required for the client/node instantiation (Section~\ref{sec:two-instantiations}). We view a full-covariance or block-diagonal FIM estimator, and an adaptive schedule for $\alpha,\gamma$, as natural directions for future work.\\

\noindent \textbf{Behavior under recombination.} Baseline BL1 (standard cross-validation on the unfragmented training set; Section~\ref{sec:expdesign}) already evaluates the recombined-data setting. As shown in Tables~\ref{tab: bst-CV-full} and \ref{tab: ccca}, BL1 consistently performs at least as well as the fragmented baseline (BL2) and is generally comparable to \texttt{PIcsC}, indicating that the proposed regularization does not degrade performance when data are considered jointly. This behavior is expected because the Fisher-based parameter prior is designed to preserve information accumulated across fragments rather than discard it (Section~\ref{sec:fim-approx}). We have not, however, performed a dedicated catastrophic-forgetting experiment in which a model trained on fragmented data is subsequently evaluated after explicit recombination. A systematic evaluation following continual-learning protocols remains an interesting direction for future work.

\noindent \textbf{Tabular versus image datasets.} Partition-induced covariate shift is more pronounced on the KEEL tabular benchmarks (Tables~\ref{tab: cccakeel}, \ref{tab: c3keel}) than on the image benchmarks, despite the simpler MLP architecture. This is primarily due to dataset size rather than model complexity. The KEEL datasets contain only hundreds to a few thousand examples, so fragmentation produces much smaller partitions, resulting in poorer estimates of the local covariate distribution and greater sampling-induced shift. This observation is consistent with the trend across Tables~\ref{tab: bst-CV-full}--\ref{tab: c3fold}, where increasing the number of partitions (and thus reducing per-partition sample support) leads to greater variance and covariate shift (Section~\ref{sec:e1e2-results}).

\section{Conclusions}
\label{5}
We propose $PIcsR$, a Fisher-information-based framework for remediating partition-induced covariate shift, whether caused by the deliberate fragmentation of a centrally held dataset into sub-datasets (batches or folds for cross-validation), or by the native non-colocation of data across clients or nodes in distributed and federated learning. The central finding of this paper is that these two settings, ordinarily studied by disjoint literatures (Section~\ref{bnchmrks}), are instances of a single phenomenon: covariate shift induced purely by how data is partitioned and exposed to a learner, independent of any shift in the underlying population. A single Fisher-Cram\'er-Rao estimator, requiring only fragment-local gradient statistics and no raw data exchange, both quantifies and corrects this shift in both settings, with only a conditional detection trigger (Section~\ref{sec:conditional}) distinguishing the bounded, all-pairs batch/fold case from the unbounded, streaming federated case. $PIcsR$ demonstrates \textit{double mitigation}, remediating both natural and partition-induced shift simultaneously, consistently across datasets and fragmentation arrangements (Sections~\ref{sec:e1e2-results}-\ref{sec:e3e4-results}), and extends without modification of its core estimator to federated deployment (Section~\ref{sec:fl-results}). $PIcsR$ therefore sits at the core of cross-validation and distributed training alike, wherever the entire training set may not be available in one place at one time.

\noindent Beyond the specific results reported here, we see three directions for future work, expanding on the limitations discussed in Section~\ref{sec:discussion}. First, a full-covariance or block-diagonal FIM estimator would relax the diagonal approximation used throughout, at increased but potentially still tractable memory cost, and may improve remediation accuracy on datasets with strongly correlated parameters. Second, an adaptive schedule for the conditional mechanism's hyperparameters ($\alpha$, $\gamma$, and $\lambda$ itself; Sections~\ref{sec:conditional}, \ref{cp}), rather than the fixed values calibrated in this paper, would reduce the need for per-deployment tuning and could improve robustness across architectures and loss functions. Third, a dedicated evaluation of forgetting under fragment recombination, and of $PIcsR$'s behavior when fragments are revisited or reordered, would clarify its relationship to the broader continual-learning literature on task-agnostic distributional shift (Section~\ref{bnchmrks}) and further test the double-mitigation property under conditions not covered by the batch/fold and federated experiments reported here.\\

\noindent
In summary, \texttt{PIcsC} consistently improves performance across both centralized and distributed learning settings affected by partition-induced covariate shift. Compared with existing state-of-the-art methods, it improves average accuracy by more than $9.5\%$ on datasets without natural covariate shift and by more than $20\%$ on datasets exhibiting natural covariate shift (Section~\ref{sota}). Under fragmented training, \texttt{PIcsC} increases average accuracy by more than $10\%$ and $25\%$ in the batchwise setting (Section~\ref{sec:e1e2-results}), and by more than $43\%$ and $28\%$ for image-based and tabular datasets, respectively, in the foldwise setting (Section~\ref{sec:e3e4-results}). The proposed Fisher-based formulation scales linearly with the size of an individual fragment  whether a batch, fold, or client shard  independent of the size of the complete dataset (Section~\ref{sec:fim-approx}). Furthermore, the proposed conditional detection-and-adaptation mechanism naturally extends the framework to client/node fragmentation, consistently outperforming FedAvg, FedProx, and SCAFFOLD by 3 -5 percentage points on seven non-IID federated learning benchmarks without requiring client-specific personalization (Section~\ref{sec:fl-results}).
\subsection*{Author Contributions}
Behraj Khan: conception and implementation of the method, experimental design and execution, manuscript writing. Behroz Mirza: refinement and analysis of experiments, manuscript writing. Tahir Qasim Syed: conception of the method, manuscript revision, and supervision. All authors reviewed and approved the final manuscript.

\bibliography{PIcsR-arXiv}

\end{document}